\documentclass[preprint,12pt]{elsarticle}

\usepackage{amsmath, amssymb}
\usepackage{array}
\usepackage{booktabs}
\usepackage[utf8]{inputenc}
\usepackage[T1]{fontenc}
\usepackage{lmodern}
\usepackage{geometry}
\usepackage{setspace}
\usepackage{graphicx}

\usepackage{makecell}
\usepackage{multirow}
\usepackage{natbib}
\usepackage{tabularx}
\usepackage{xcolor}

\usepackage{lineno}

\usepackage{hyperref}

\journal{Computerized Medical Imaging and Graphics}

\begin{document}
\nolinenumbers
\begin{frontmatter}

\title{Benchmarking Intra-Patient 3D Deformable Multimodal Image Registration} 

\author[imag2]{Matteo BARBIERI} 
\author[replico,imag2]{Giammarco LA BARBERA} 
\author[replico,imag2]{Juan Pablo DE LA PLATA} 
\author[pediatric,imag2]{Sabine SARNACKI} 
\author[sorbonne,imag2,ltci]{Isabelle BLOCH} 
\author[ltci]{Pietro GORI\corref{cor1}} 

\cortext[cor1]{Corresponding author} 
\ead{pietro.gori@telecom-paris.fr} 

\affiliation[imag2]{ 
    organization={IMAG2, Institut IMAGINE, Université Paris Cité}, 
    city={Paris}, 
    country={France} 
}

\affiliation[sorbonne]{ 
    organization={Sorbonne Université, CNRS, LIP6}, 
    city={Paris}, 
    country={France} 
} 

\affiliation[ltci]{ 
    organization={LTCI, Télécom Paris, Institut Polytechnique de Paris}, 
    city={Palaiseau},
    country={France} 
}

\affiliation[replico]{ 
    organization={Replico SAS}, 
    city={Paris}, 
    country={France} 
} 

\affiliation[pediatric]{ 
    organization={Université Paris Cité, Department of Pediatric Surgery, Hôpital Necker Enfants-Malades, Assistance Publique--Hôpitaux de Paris (AP-HP)}, 
    city={Paris}, 
    country={France} 
}

\begin{abstract} 
Multimodal image registration is a key component of many clinical workflows, yet it remains challenging because corresponding anatomical structures often exhibit substantially different image intensities across modalities. In this work, we present a comprehensive benchmark of intra-patient 3D multimodal deformable registration methods across three datasets covering different anatomical regions and difficulty levels, including both synthetic deformation recovery and real clinical scenarios.

We evaluate classical optimization-based approaches and modern learning-based methods, including recent deep learning and foundation models, using complementary metrics: Average Dice similarity coefficient ($\overline{\mathrm{DSC}}$), average 95\textsuperscript{th}-percentile Hausdorff distance ($\overline{\mathrm{HD95}}$), and a modality-independent structural similarity measure based on the MIND self-similarity context (MIND-SSC). Results show high variability across datasets, with learning-based methods demonstrating superior performance on large synthetic benchmarks, while only limited improvements are observed in real pelvic registration.

A key finding of this study is the consistent disagreement between geometric metrics ($\overline{\mathrm{DSC}}$, $\overline{\mathrm{HD95}}$) and image-based similarity metrics (MIND-SSC), highlighting that improved overlap does not necessarily imply better global multimodal correspondence. Furthermore, anatomy-guided approaches achieve the highest overlap scores but exhibit degraded performance outside of segmented regions, revealing a trade-off between label-driven alignment and global structural coherence. 

Overall, our results indicate that no current method achieves robust performance across anatomies and modalities. We demonstrate that intra-patient 3D multimodal registration requires multi-criteria evaluation, including deformation-based metrics, and remains an open problem. This benchmark provides guidance for future research toward more generalizable, anatomically consistent, and clinically relevant registration frameworks. 
\end{abstract}


\end{frontmatter}


\section{Introduction}
Image registration is one of the building blocks for achieving personalized precision medicine, as it enables the consistent spatial alignment of anatomical structures across images acquired using different modalities, at different time points or from different patients. By establishing spatial correspondences, registration plays a critical role in numerous clinical and research applications, including longitudinal disease monitoring \cite{durrleman2009spatiotemporal,lorenzi2013efficient}, population analysis \cite{gori2013bayesian,gori2017bayesian}, image-guided interventions \cite{zanello2021automated}, as well as fundamental research in the understanding of the human organs, such as the brain, where complementary anatomical information and functional information are integrated into a common reference frame~\cite{2013Sotiras, 2021Xiao,2022Zou}.

Formally, image registration consists in estimating an optimal spatial transformation ---  typically represented as a dense displacement field or a parametric mapping --- that aligns a moving image onto a fixed reference image. Transformation models are commonly divided into linear (e.g., rigid or affine), which apply global transformations governed by a small number of parameters, and deformable and non-linear models, which allow for spatially varying deformations and are essential for capturing complex anatomical variability~\cite{2019DeVos}. Deformable registration is often further constrained through regularization terms to ensure smoothness, topology preservation, or diffeomorphic properties, thereby enforcing anatomically plausible mappings~\cite{2008Avants,2019Dalca}.

From an algorithmic perspective, registration approaches are traditionally classified into classical iterative methods and learning‑based (data‑driven) methods. Classical methods formulate registration as an energy minimization problem combining a similarity measure and a regularization term, solved iteratively using numerical optimization techniques. These approaches benefit from well-established theoretical foundations and explicit control over transformation properties~\cite{2013Sotiras}. By contrast, learning-based methods aim to directly infer transformation using neural networks, trading explicit modeling for statistical learning and enabling real-time inference once trained~\cite{2025Chenb, 2022Zou}.

In addition to these general categorizations, several application-dependent factors critically influence the choice of a registration strategy. Among them, the imaging modalities of the input images (e.g., Computed Tomography (CT), Magnetic Resonance Imaging (MRI), Ultrasound (US)) and whether they correspond to the same modality (monomodal) or different modalities (multimodal) is particularly important, as it determines the validity of assumptions regarding intensity correspondence. Other factors include the dimensionality of the data (e.g. 2D vs. 3D), the anatomical region of interest (e.g. brain, pelvis, abdomen, etc.), and whether the task involves intra-subject (longitudinal) or inter-subject (population-level) alignment~\cite{2020Fu,gori2015joint,8307447}. These considerations directly drive the design of similarity measures \cite{charon2019fidelity}, transformation models, and learning strategies.

Since the advent of digital medical imaging, research has focused mainly on monomodal registration, achieving significant improvement of similarity and structure-based metrics even in cases with complex geometrical mismatch. Multimodal registration, used for multiparametric studies or intra-operative visualization, still represents a challenging problem, as the assessment of registration quality through superimposition of anatomical features can no longer pass through intensity-based metrics. While some authors have attempted to provide better suited similarity metrics such as MIND~\cite{2012Heinrich} or MIND-SSC~\cite{2013Heinrich}, others have developed learning-based similarity approximations such as REE-UNET~\cite{2026Nascimento} and used them in registration pipelines~\cite{2025bHe, 2024Mok, 2025Tursynbek}.

This work specifically focuses on deformable intra-patient multimodal registration. Our aim is to provide a methodological and clinically-relevant benchmark on two publicly available datasets and one private dataset for a selected number of state-of-the-art multimodal deformable registration strategies. This benchmark evaluates two distinct capabilities: (i) recovery of known transformations under modality shifts, and (ii) estimation of true correspondence in realistic multimodal settings. Based on synthetic and real misaligned data, we perform a comparative study of the registration ability of different methods in setups of increasing difficulty, while varying modalities and regions of interest. Eventually, we evaluate the current state of multimodal deformable registration and provide the research community with guidelines for further work.

The paper is organized as follows. After summarizing related work in Section~\ref{sec:relatedWork}, a brief introduction to the functioning of each evaluated method is provided in Section~\ref{sec:evaluated-registration-methods}. The specifics of the benchmarking methodology --- such as datasets, protocols and evaluation metrics --- are described in Section~\ref{sec:Methods}, and associated results are presented in Section~\ref{sec:Results}. Eventually, a discussion presenting the outcomes and limitations of the current benchmark, as well as guidelines for future work on deformable multimodal image registration is provided in Section~\ref{sec:Discussion}. A conclusion summarizing the findings is provided in Section~\ref{sec:Conclusion}.


\section{Related work}
\label{sec:relatedWork}

Let \( I_T, I_M : \Omega \to \mathbb{R} \) denote the target (i.e. reference) and moving images, respectively, defined on a domain of the $d$-dimensional real space ($\Omega \subset \mathbb{R}^d$). Deformable image registration aims to estimate a spatial transformation \( \phi : \Omega \to \Omega \) such that \( I_M \circ \phi\approx~I_T\), so that the moving image $I_M$, warped by the transformation $\phi$, should match the target image $I_T$. This problem is commonly formulated as the minimization of an energy functional of the following form:
\begin{equation}\label{eqn:energy-functional}
    \mathcal{E}(\phi) = \mathcal{D}\big(I_T, I_M \circ \phi) + \lambda \, 
    \mathcal{R}(\phi),
\end{equation}
where \( \mathcal{D} \) is a similarity measure between images, \( \mathcal{R} \) is a regularization functional enforcing smoothness or physical plausibility, and \( \lambda~>~0 \) balances the two terms.

Different registration methods can be distinguished by their parameterization of $\phi$, the choice of similarity metric $\mathcal{D}$, and the form of regularization $\mathcal{R}$. To better understand the difficulties of multimodal registration, and since some monomodal registration methods are sometimes directly used for multimodal registration, we present in the following section registration methods for both the monomodal and multimodal registration problem. Please note that we will consider only methods accounting for geometric transformations, without intensity/iconographic modifications, thereby excluding frameworks such as metamorphosis \cite{francois2021metamorphic,francois2022weighted,maillard2022deep}.

\subsection{Monomodal image registration}
Monomodal registration is characterized by the assumption that voxels representing the same point in the physical space within moving and target images have similar intensity distributions. In this setting, intensity-based similarity metrics such as sum of squared differences (SSD) or cross-correlation (CC) are well suited, enabling accurate alignment in applications such as longitudinal analysis, where one can assume that moving $I_M$ and target $I_T$ images only differ by Gaussian noise (SSD) or that there is a linear relationship between them (CC).

Among the classical approaches, Advanced Normalization Tools (ANTs)~\cite{2008Avants} and Elastix~\cite{2009Klein} have emerged as widely adopted frameworks for medical image registration. In particular, the Symmetric Normalization (SyN) algorithm~\cite{2008Avants}, based on a diffeomorphic formulation with a symmetric energy functional, is often regarded as the reference method for brain MRI registration due to its robustness, inverse consistency, and strong performance in benchmark evaluations. For applications involving large, spatially heterogeneous, and highly non-linear deformations (such as those encountered in thoracic or abdominal imaging), dense non-parametric approaches remain essential. In particular, free-form deformation (FFD) models based on B-splines~\cite{2002Rueckert} provide flexible local control, while variational diffeomorphic frameworks such as Large Deformation Diffeomorphic Metric Mapping (LDDMM)~\cite{gori2017bayesian, ashburner2011diffeomorphic, avants2008symmetric, 2005Beg,charon2019fidelity} offer strong theoretical guarantees on smoothness and invertibility. In parallel, algorithms based on optical-flow inspired updates such as Deformable demons and its diffeomorphic extensions remain highly competitive due to their computational efficiency and ability to handle large displacement fields~\cite{2014Lombaert,2009Vercauteren}.

More recently, deep learning-based methods have significantly transformed monomodal registration by replacing iterative optimization with feedforward inference. Early frameworks such as VoxelMorph~\cite{2019Balakrishnan} demonstrated that Convolutional Neural Networks (CNN) can learn to predict dense deformation fields in an unsupervised manner using differentiable spatial transformers. Subsequent extensions introduced probabilistic formulations and diffeomorphic constraints via stationary velocity fields~\cite{2019Dalca}, improving both robustness and topology preservation. Alternatively, methods such as GradICON~\cite{2023Tian} enforce inverse consistency through gradient-based regularization, leading to improved numerical stability and more consistent mappings. Meanwhile weakly-supervised frameworks incorporate anatomical priors by means of keypoints or full organ segmentations to enforce reliable anatomical displacement fields~\cite{2018Hu, 2022Jian}. A significant difference between the developments of deep learning methods with respect to the previously described classical methods is that the expression of physical properties (such as diffeomorphism) is not formally enforced but rather implicitly encouraged through the tuning of parameters that balance competing criteria.

Building on these convolutional approaches, more recent architectures leverage transformer-based designs to better capture long-range spatial dependencies. Unlike CNNs, which rely on local receptive fields and hierarchical feature aggregation, transformers exploit global context through self-attention mechanisms, enabling direct interactions between distant regions of the image. This property is particularly advantageous in medical image registration, where large anatomical displacements and weak local correspondences can challenge purely local models. Transformer-based registration methods such as TransMorph~\cite{2022Chen}, Vit-v-Net~\cite{2021bChen} and others reported by Ramadan \textit{et al.} in~\cite{2024Ramadan} typically combine attention modules with multi-scale representations, allowing them to jointly encode fine-grained details and global structural alignment.

Finally, hybrid methods have emerged to bridge the gap between purely data-driven approaches and classical variational frameworks. These approaches integrate learned components, such as similarity metrics or feature embeddings, into explicit energy minimization schemes. For example, learned similarity metrics have been shown to outperform handcrafted metrics in challenging registration scenarios~\cite{2019Haskins, 2019Niethammer, 2016Simonovsky}. More recently, learned semantic feature representations have been incorporated into optimization-based registration frameworks to improve robustness across modalities while retaining the interpretability and regularization properties of classical formulations~\cite{2021Czolbe,2020Haskins}.

\subsection{Multimodal image registration}
By contrast, multimodal image registration remains significantly more challenging due to the absence of direct intensity correspondence between modalities such as Computed Tomography (CT), Magnetic Resonance Imaging (MRI) or ultrasound. Early approaches addressed this issue using similarity metrics derived from information theory, most notably Mutual Information (MI)~\cite{2002Maes}, which quantifies statistical dependence between image intensities and remains a strong baseline for both rigid and deformable registration, alongside early gradient-based formulations~\cite{2006Haber}. However, MI suffers from limited spatial specificity and sensitivity to interpolation artifacts. To overcome these limitations, a second class of methods introduced modality-invariant descriptors that rely on structural (rather than intensity-based) information. A prominent example is the Modality Independent Neighborhood Descriptor (MIND-SSC)~\cite{2013Heinrich}, which encodes local self-similarity patterns and has become a standard for multimodal deformable registration. Other descriptors, such as binary gradient angle representations~\cite{2017Jiang} or self-similarity-based features~\cite{2013Heinrich,2023Wang}, further exploit the structural consistency of anatomical patterns across modalities.

Deep learning has had a particularly strong impact in the multimodal setting, as it enables the learning of modality-invariant representations directly from data. Early works explored weakly-supervised and adversarial learning strategies~\cite{2019Fan,2018Hu,2020Xu}, as well as deep similarity metrics~\cite{2019Haskins}, to bridge modality gaps. Other approaches relied on image-to-image translation~\cite{2021Kim,2021Kong,2019Wei}, although their effectiveness has been shown to be modality-dependent and sometimes inferior to direct registration methods~\cite{2021Lu}. More recent methods leverage contrastive learning and shared embedding spaces~\cite{2024Mok, 2022Dey,2020Pielawski}, cycle-consistent training strategies~\cite{2021Kim}, cross-modal attention mechanisms~\cite{2022Song}, or self-supervised anatomical representations~\cite{2021Liu}. In parallel, differentiable similarity learning frameworks such as DISA~\cite{2023Ronchetti} have been proposed to generalize similarity modeling across modalities.

Contemporary state-of-the-art approaches focus on end-to-end deformation prediction with strong geometric constraints. Examples include reinforcement learning-based formulations~\cite{2021Hu}, geometry-consistent adversarial models~\cite{2023Liu}, and transformer-based architectures~\cite{2022Chen}. Recent advances also explore foundation models and large-scale pretraining, such as multiGradICON~\cite{2024Demir}, which extends inverse-consistent diffeomorphic frameworks to multimodal settings, and universal matching models~\cite{2025He}. Additional recent contributions include anatomy-aware architectures such as MAIRNet~\cite{2024Gao} and strategies that explicitly reduce modality gaps via mono-modalization~\cite{2025Choo}. Emerging directions further include diffusion-based similarity modeling~\cite{2025Tursynbek}, implicit neural representations~\cite{2025Wang}, and modality-agnostic feature learning from large pretrained models~\cite{2025bHe}.

\subsection{Limitations of existing benchmarks and methods}
Overall, while classical methods based on MI and handcrafted descriptors remain competitive in controlled or low-variability scenarios, learning-based approaches increasingly define the state of the art in multimodal deformable registration, particularly in settings involving large modality gaps, heterogeneous datasets, and real-time constraints. The field is progressively moving toward unified, generalizable frameworks capable of handling diverse modalities, anatomies, and acquisition conditions within a single model.

However, the effectiveness of both classical and learning-based multimodal deformable registration methods remains limited in many clinical applications where sub-millimetric to millimetric accuracy (typically $<1$--2~mm) is required, such as stereotactic radiotherapy, neurosurgical navigation, and image-guided interventions. Despite substantial progress, extensive evaluations consistently report residual Target Registration Errors (TRE) in the range of 1--3~mm for state-of-the-art deformable registration methods, even in relatively controlled settings such as the Learn2Reg challenge \cite{2022Hering}. The Learn2Reg challenge editions (2020--2024) provide a comprehensive multi-anatomy benchmarking framework that reveals persistent task-dependent gaps~\cite{2025Hansen, 2022Hering}: median TRE values of approximately 1--2~mm are achievable in brain registration, whereas abdominal registration remains substantially harder, with 95\textsuperscript{th}-percentile Hausdorff distances reaching up to 20~mm even for top-ranked methods. These results highlight the impact of respiratory motion, sliding organ interfaces, and large non-linear deformations that introduce fundamental ambiguities in defining voxel-level correspondence, particularly in homogeneous or biomechanically complex regions~\cite{2013Sotiras, 2006Crum}. Furthermore, among the best-performing methods in Learn2Reg 2024, intra-method variability in TRE can exceed 1~mm across tasks, underscoring that no single architecture generalizes robustly across anatomies and modalities~\cite{2025Hansen}.

In the case of larger structures, the lack of quality is also visible from the Dice results. For instance, on large-scale brain MRI benchmarks (e.g. OASIS, ABIDE, ADHD200), mean Dice Similarity Coefficient (DSC) scores for cortical and subcortical structures cluster around 75--78\% for modern approaches such as VoxelMorph and its diffeomorphic variants~\cite{2019Dalca, 2019Balakrishnan}.

The challenge becomes even more pronounced in scenarios involving highly heterogeneous image appearances, large unconstrained fields of view, or significant modality-specific artifacts~--- such as speckle noise in ultrasound or bias fields in MRI. In registration of prostate MRI with Transrectal Ultrasound (TRUS), one of the most clinically relevant and extensively studied multimodal tasks, state-of-the-art methods~--- including learning-based approaches --- consistently report TRE values in the range of 4--11~mm, with only marginal improvements over classical iterative techniques despite the focus on small regions of interest during prediction~\cite{2019Haskins}. 
This is particularly concerning, given that the clinically tolerated targeting error for prostate biopsy guidance is often cited as $\leq$2.5~mm~\cite{2015Tilak}.
More broadly, recent surveys confirm that ultrasound-based multimodal registration constitutes one of the most challenging subfields, reporting an average TRE of 4.95~mm for the best considered method on a manually curated dataset~\cite{2023Wang}. Even methods specifically designed for CT--ultrasound or MRI--ultrasound settings, including those leveraging self-similarity descriptors~\cite{2012Heinrich,2013Heinrich} or cross-modal attention~\cite{2022Song}, have not demonstrated consistent sub-3~mm accuracy in prospective clinical evaluations, reflecting the fundamental difficulty of establishing reliable correspondences across such dissimilar intensity distributions.

In addition, the performance and generalization of deep learning-based registration methods are strongly constrained by the scarcity of large-scale annotated datasets with accurate voxel-wise correspondence ground truth. Unlike for the monomodal settings, multimodal registration inherently lacks ``ground truth'' deformation fields, necessitating weakly-supervised, synthetic, or self-supervised training strategies that introduce additional uncertainty~\cite{2018Hu,2022Zou}. This issue is further compounded by domain shift across scanners, acquisition protocols, and patient populations, which have been shown to degrade TRE performance by several millimeters when models are evaluated outside their training distribution~\cite{2022Hering}. Concretely, methods achieving median TRE values below 2~mm on in-distribution test sets have been observed to exceed 4--5~mm on out-of-distribution data~\cite{2025Hansen,2022Hering}, a degradation that is especially acute for learning-based methods that overfit to scanner-specific contrast profiles or resolution characteristics. 
Recent large-scale approaches, including inverse-consistent diffeomorphic frameworks such as multiGradICON~\cite{2024Demir}, aim to address these challenges through improved regularization, symmetry constraints, and scalability to diverse training distributions. While these methods demonstrate improved stability and competitive benchmark performance, evaluations on population-scale datasets, such as UK Biobank~\cite{2015Sudlow}, reveal persistent variability in alignment quality, with residual errors still exceeding clinically desirable thresholds (Dice coefficient > 0.8, TRE > 5~mm)~\cite{2017Brock} in a non-negligible fraction of cases, highlighting the gap between controlled benchmark performance and real-world clinical deployment.

\section{Evaluated registration methods}
\label{sec:evaluated-registration-methods}
Based on the previous analysis of related work, we focus on largely used, state-of-the-art methods. For better representativeness, we evaluate methods spanning classical optimization, handcrafted multimodal descriptors, and modern deep learning approaches including CNN- and foundation-model-based registration frameworks. The methods were selected on the basis of how widely they were used and the availability of ready-to-use packages or github implementations.

Given these constraints, Table~\ref{tab:Methods-and-implementations} summarizes the methods investigated in this study, reporting their original publications and how to retrieve them, while a brief presentation of each of them is provided below. Given the common trend of using handcrafted similarity-based metrics when training DL methods, we chose to use the MIND-SSC loss metric ($\mathcal{L}_{MIND-SSC}$) as a common ground when training models. This also includes the pretrained version of the multiGradICON foundation model, which was already trained using MIND-SSC loss by the authors~\cite{2024Demir}. 

\paragraph{Anatomically-guided registration}
\label{par:anatomically-guided-registration} 
In addition to the proposed experiments, an assessment of the performance of anatomically-guided registration is proposed and implemented for the ANTs SyN and GradICON methods, respectively called ANTS SyN seg. and GradICON seg. in the following, representing the classical and deep-learning side of the optimization strategies. For classical methods, such as ANTs SyN, this is performed by giving as input the segmentation masks of homologous structures in $I_M$ and $I_T$ instead of the target and moving images. The resulting deformation map is then applied to the full image. For GradICON, the modified implementation consists in the introduction of an additional soft Dice loss term, evaluating the matching of the registered (linear interpolation) and target segmentations, thus favoring the alignment of segmented structures during training.

\begin{table}[htbp]
    \centering
    \caption{Considered multimodal deformable registration methods for benchmark.}
    \label{tab:Methods-and-implementations}
    \begin{tabularx}{\textwidth}{l X l X}
        \toprule
        \textbf{Method} & \textbf{Reference publication} & \textbf{Category} & \textbf{Code repository} \\
        \midrule
        ANTs SyN         & \cite{2008Avants}  & Classic & \url{https://antspyx.readthedocs.io/en/latest/} \\
        Elastix          & \cite{2009Klein}   & Classic & \url{https://github.com/InsightSoftwareConsortium/ITKElastix} \\
        Deforming demons & \cite{2009Vercauteren, 2014Lombaert}  & Classic & \url{https://github.com/InsightSoftwareConsortium/ITKElastix} \\
        VoxelMorph       & \cite{2019Balakrishnan}  & CNN & \url{https://github.com/htwin/voxelmorph_torch/tree/master} \\
        GradICON         & \cite{2023Tian}  & CNN & \url{https://github.com/MICV-yonsei/M2M-Reg} \\
        M2M-GradICON     & \cite{2025Choo}  & CNN & \url{https://github.com/MICV-yonsei/M2M-Reg} \\
        multiGradICON    & \cite{2024Demir} & Foundation model & \url{https://github.com/uncbiag/uniGradICON} \\
        \bottomrule
    \end{tabularx}
\end{table}

\paragraph{ANTs SyN}
ANTs SyN~\cite{2008Avants} performs deformable registration in the space of diffeomorphisms, ensuring smoothness and invertibility of the estimated mappings. The transformations are generated by time-dependent velocity fields through forward $\phi_{MT}$ and backward $\phi_{TM}$ flows, defined by:

\begin{equation} 
    \begin{aligned} 
        \frac{\partial \phi_{MT}(x,t)}{\partial t} 
        &= v_{MT}(\phi_{MT}(x,t),t), 
        &\qquad \phi_{MT}(x,0) &= x, \\ 
        \frac{\partial \phi_{TM}(x,t)}{\partial t} 
        &= v_{TM}(\phi_{TM}(x,t),t), 
        &\qquad \phi_{TM}(x,0) &= x. 
    \end{aligned} 
\end{equation}

Here, $\phi_{MT}(x,t)$ (resp. $\phi_{TM}(x, t)$) denotes the position at time $t$ of a point initially located at $x$, i.e., a pointwise evaluation of the flow, whereas $\phi_{MT}(\cdot,t)$ (resp. $\phi_{TM}(\cdot,t)$) denotes the full deformation map at time $t$, i.e., a diffeomorphism from $\Omega$ to $\Omega$.

SyN adopts a symmetric formulation by introducing two velocity fields $v_{MT}(x,t)$ and $v_{TM}(x,t)$, defined for $t \in [0,\tfrac{1}{2}]$, which transport the moving image $I_M$ and the target image $I_T$ toward a common midpoint. The corresponding energy functional is
\begin{equation}
    \mathcal{E}(v_{MT},v_{TM}) =
    \int_0^{1/2} \left( \|L v_{MT}(\cdot,t)\|_2^2 + \|L v_{TM}(\cdot,t)\|_2^2 \right)\, dt
    + \mathcal{D}\Big( I_M \circ \phi_{MT}(\cdot,\tfrac{1}{2}),\,
                       I_T \circ \phi_{TM}(\cdot,\tfrac{1}{2}) \Big),
\end{equation}
where $L$ is a differential operator enforcing regularity and $\mathcal{D}$ is a similarity measure between images. The final transformation mapping $I_M$ to $I_T$ is obtained as
\begin{equation}
    \phi = \phi_{MT}(\cdot,\tfrac{1}{2}) \circ \phi_{TM}^{-1}(\cdot,\tfrac{1}{2}).
\end{equation}
This symmetric construction makes the method invariant to the choice of reference image and improves inverse consistency.

\paragraph{Deforming Demons}
Deforming Demons estimates a dense displacement field $\phi_{MT}$ through iterative updates derived from optical flow principles. At each iteration, the update is computed as
\begin{equation}
    \phi_{MT}^{k+1} = \phi_{MT}^k + \frac{(I_T - I_M \circ (Id + \phi_{MT}^k)) \nabla I_T}{(I_T - I_M \circ (Id + \phi_{MT}^k))^2 + \|\nabla I_T\|_2^2},
\end{equation}
which aligns image intensities using local gradient information. The displacement field is regularized via Gaussian smoothing to enforce spatial coherence. Extensions to diffeomorphic Demons introduce symmetrization or parametrize the transformation by a stationary velocity field $v_{MT}$, and compute $\phi_{MT} = \exp(v_{MT})$, ensuring invertibility while retaining the same local update mechanism.

\paragraph{Elastix parametric registration}
Elastix implements parametric registration by modeling the transformation as a linear combination of basis functions. The deformation is expressed as
\begin{equation}
    \phi_{MT}^\theta(x) = x + \sum_{i=1}^p \theta_i \beta_i(x),
\end{equation}
where $\beta_i$ are typically B-spline basis functions defined on a control point grid, and $p$ is the number of such functions. The parameters $\theta_i$ are optimized by minimizing the energy functional given in Equation~\ref{eqn:energy-functional}. This formulation constrains the deformation to a low-dimensional space, enabling efficient optimization and explicit control over smoothness.

\paragraph{VoxelMorph}
VoxelMorph formulates registration as a supervised or unsupervised learning problem, where a neural network $f_\theta$ predicts a dense deformation field from image pairs:
\begin{equation}
    \phi_{MT} = f_\theta(I_T, I_M).
\end{equation}
The network is trained by minimizing the energy functional of Equation~\ref{eqn:energy-functional} with $\mathcal{R} = \|\nabla \phi_{MT}\|_2^2$. Diffeomorphic variants predict a stationary velocity field $v_{MT}$ and compute $\phi_{MT} = \exp(v_{MT})$ using scaling-and-squaring integration, ensuring invertibility of the transformation.

\paragraph{GradICON}
GradICON learns bidirectional transformations between image pairs and enforces inverse consistency during training. The model predicts forward and backward deformations $\phi_{TM}$ and $\phi_{MT}$, and minimizes a loss function that includes the inverse-consistency constraint $\mathcal{L}_{IC}$
\begin{equation}
    \mathcal{L}_{IC} = \|\nabla[\phi_{TM} \circ \phi_{MT}] - Id\|_2^2,
\end{equation}
as an additional regularizer $\mathcal{R}$, which penalizes deviations from perfect inversion. The full objective function also includes an image similarity term and smoothness regularization. Spatial image gradients are incorporated in the learning process to stabilize correspondence estimation and improve alignment near edges.

\paragraph{M2M-GradICON}
\begin{figure}[ht!]
    \centering
    \includegraphics[width=0.6\textwidth]{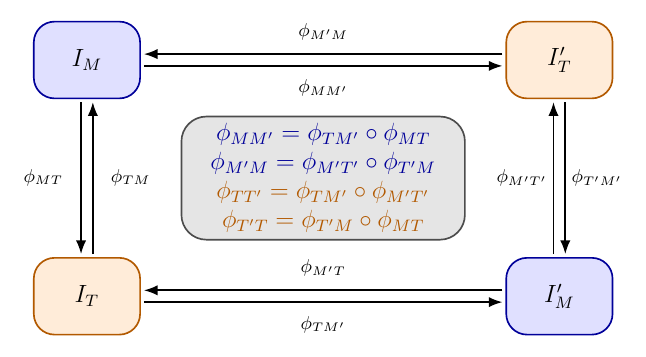}
    \caption{Schematic cyclic registration from M2M-GradICON.}
    \label{fig:M2M-scheme}
\end{figure}
M2M-GradICON addresses multimodal registration by first mapping the input images into a shared, mono-modal representation. To enforce consistency between the two modalities, the authors introduce a four-image cyclic registration scheme, illustrated in Figure~\ref{fig:M2M-scheme}. Specifically, in addition to the original reference and moving images, $I_T$ and $I_M$, the authors also consider their corresponding (\textit{bridge}) images in the shared representation, $I_{T'}$ and $I_{M'}$. Deformation fields are estimated between the original and shared representations, namely $\phi_{MM'}$ and $\phi_{M'M}$ between $I_M$ and $I_{M'}$, and $\phi_{TT'}$ and $\phi_{T'T}$ between $I_T$ and $I_{T'}$. Two weighting hyperparameters, $\lambda_{can}$ and $\lambda_{IC}$, control the contributions of the canonical consistency regularizer $\mathcal{L}_{can}$ and the inverse-consistency regularizer $\mathcal{L}_{IC}$, respectively. The resulting cost is :
\begin{equation}
    \mathcal{L} = \mathcal{L}_{sim} + \lambda_{can}\mathcal{L}_{can} + \lambda_{IC}\mathcal{L}_{IC},
\end{equation}
with 
\begin{align*}
    \mathcal{L}_{sim} =&\mathcal{D}\!\big((I_M \circ \phi_{MM'}), I_{M'}\big)
                        +
                        \mathcal{D}\!\big((I_{M'} \circ \phi_{M'M}), I_{M}\big)+\\
                        &\mathcal{D}\!\big((I_T \circ \phi_{TT'}), I_{T'}\big)
                        +
                        \mathcal{D}\!\big((I_{T'} \circ \phi_{T'T}), I_{T}\big)
\end{align*}
\begin{align*}
    \mathcal{L}_{can} &= \|\nabla[\phi_{MT}\circ\phi_{TM'}\circ\phi_{M'T'}\circ\phi_{T'M}-Id]\|_2^2 \\
    \mathcal{L}_{IC} &= \|\nabla[\phi_{TM} \circ \phi_{MT}] - Id\|_2^2
\end{align*}

This ensures transitive consistency of the deformation fields. The optimization is performed over a shared network that produces pairwise mappings, resulting in a globally coherent transformation space across the dataset.

\emph{Note:} The present description matches the code provided by the github link in Table~\ref{tab:Methods-and-implementations}, which is slightly different from the one described in~\cite{2025Choo}.

\paragraph{multiGradICON}
multiGradICON incorporates multi-scale and group-wise registration within an inverse-consistent learning framework. The deformation is represented as a composition of transformations across resolution levels:
\begin{equation}
    \phi = \phi^{(1)} \circ \cdots \circ \phi^{(L)},
\end{equation}
where each $\phi^{(l)}$ captures deformations at a specific scale. Training enforces similarity, smoothness, and consistency constraints at each level, improving robustness to large deformations and reducing local minima. The hierarchical structure enables coarse-to-fine alignment while maintaining global coherence across multiple images.

\section{Benchmarking methodology}
\label{sec:Methods}
This section describes the experimental framework used to evaluate and compare multimodal deformable registration methods. The benchmark is designed to assess performance across multiple anatomical regions, imaging modalities, and registration scenarios. We first present the datasets and the synthetic deformation protocol used to establish quantitative ground truth. We then describe the training strategy, evaluation metrics, and benchmarking procedures adopted to ensure a fair and reproducible comparison between classical and learning-based approaches.

\subsection{Datasets}
The benchmark study is conducted on three datasets that cover healthy and pathological subjects on different anatomical regions and multimodal configurations. The first dataset is the publicly available BRATS dataset~\cite{ 2023Adewole,2014Menze}, containing \underline{pre-registered} T1 and T2-weighted images of patients with gliomas, including low-grade and high-grade brain tumors (e.g., astrocytomas and glioblastomas). The second dataset is the publicly available Human Connectome Project (HCP)~\cite{2013VanEssen}, already pre-processed as described in~\cite{2013Glasser} and containing \underline{pre-registered} T2-weighted and diffusion $b_0$ images of healthy young adults with no major neurological abnormalities affecting brain anatomy. The third dataset, called IMAG2 (Pelvis), is a private one containing multimodal pelvic MRI of 80 pediatric subjects from the Hôpital Necker Enfants-Malades, acquired under the ClinicalTrials ID NCT06224985. It contains a mixture of control patients (10\%) and patients with pelvic tumors (40\%) or pelvic malformations (50\%). More information about this dataset can be found in \cite{pio2026imaging}.
For each patient, it contains \underline{unregistered} coronal T2-weighted MRI (reconstruction voxel size 0.86$\times$0.88$\times$0.86 $mm^3$) and multi-gradient axial DWI image (acquisition voxel size 3.3$\times$2.3$\times$3.6 $mm^3$, NEX 2, 1 $b_0$ image, 25 directions at b-value 600 $s/mm^2$) acquired on a 3T GE Discovery MR700 during the same procedure. Diffusion-weighted images were preprocessed using MRtrix3~\cite{2019Tournier}, including denoising, Gibbs ringing artifact removal, motion and eddy-current correction, and bias field homogenization. The $b_0$ image was subsequently generated by extracting the only non-diffusion-weighted (b-value = 0 $s/mm^2$) volume from the preprocessed diffusion dataset.

These datasets include brain and pelvic anatomies of control and tumor/malfomation subjects, enabling evaluation across distinct anatomical structures and deformation regimes ranging from controlled synthetic transformations to real intra-patient multimodal misalignment. All datasets are preprocessed using a standardized pipeline. For the BRATS and HCP datasets, the images are exploited at their respective resampled resolution of $1\times1\times1$~mm and $1.25\times1.25\times1.25$~mm, while the IMAG2 (pelvis) data uses a resampling of $1.25\times1.25\times3.5$~mm corresponding to the mean resolution of all available images. Intensity normalization is performed using z-scoring, and a cropping to a common Field-Of-View (FOV) is performed using the information present in the DICOM headers. Each dataset is exploited at a specific cropping size to balance computational efficiency and FOV: BRATS and HCP are cropped at $192\times192\times192$ to minimize the impact of dark background on the computation cost while the IMAG2 (pelvis) dataset is of size $192\times256\times128$, to provide a trade-off between keeping additional anatomical information and computation time. Segmentation labels for the BRATS and HCP datasets are provided by the respective project leaders. For the IMAG2 (pelvis) dataset, manual T2w image segmentations of the Hipbones, Sacrum, L5 vertebra and Bladder were provided by a medical expert and checked by at least two other experts, including an experienced radiologist. Given the difficulty of segmenting bone structures in $b_0$ images, the segmentation was obtained under the assumption that bones are rigid structures and do not change between the consecutive T2w and $b_0$ images. Hence, a rigid registration from the T2w to the $b_0$ image was performed and checked manually to align the T2w mask with visible anatomy in the $b_0$ image. The bladder was very visible in the $b_0$ and manually segmented from scratch.

For each dataset, 20\% of the subjects are held out as a test set, while maintaining the proportions of healthy vs. pathological cases as close as possible to the one of the entire set. The remaining 80\% are split into training and validation subsets using an 80/20 ratio, resulting in 64\% training, 16\% validation, and 20\% testing. Learning-based methods use the validation set to select the best model during the training process based on the minimum validation loss value, whereas classical registration methods use it for grid-search of optimal parameters. The test set is strictly held out and used exclusively for final evaluation across all methods. The number of subjects and split distribution are reported in Table~\ref{tab:Datasets-description}.

\begin{table}[htbp]
    \centering
    \caption{Dataset description.}
    \label{tab:Datasets-description}
    \begin{tabularx}{\textwidth}{l X l X}
        \toprule
        \textbf{Dataset} & \textbf{\# Cases} & \textbf{\# Train-val-test} & \textbf{Availability} \\
        \midrule
        BRATS \cite{2014Menze}  & 1251  & 801-200-250 & Public \\
        HCP \cite{2013VanEssen} & 708   & 454-113-141 & Public\\
        IMAG2 (pelvis)          & 80    & 52-12-16 & Proprietary \\
        \bottomrule
    \end{tabularx}
\end{table}

\subsection{Synthetic Deformation Protocol}
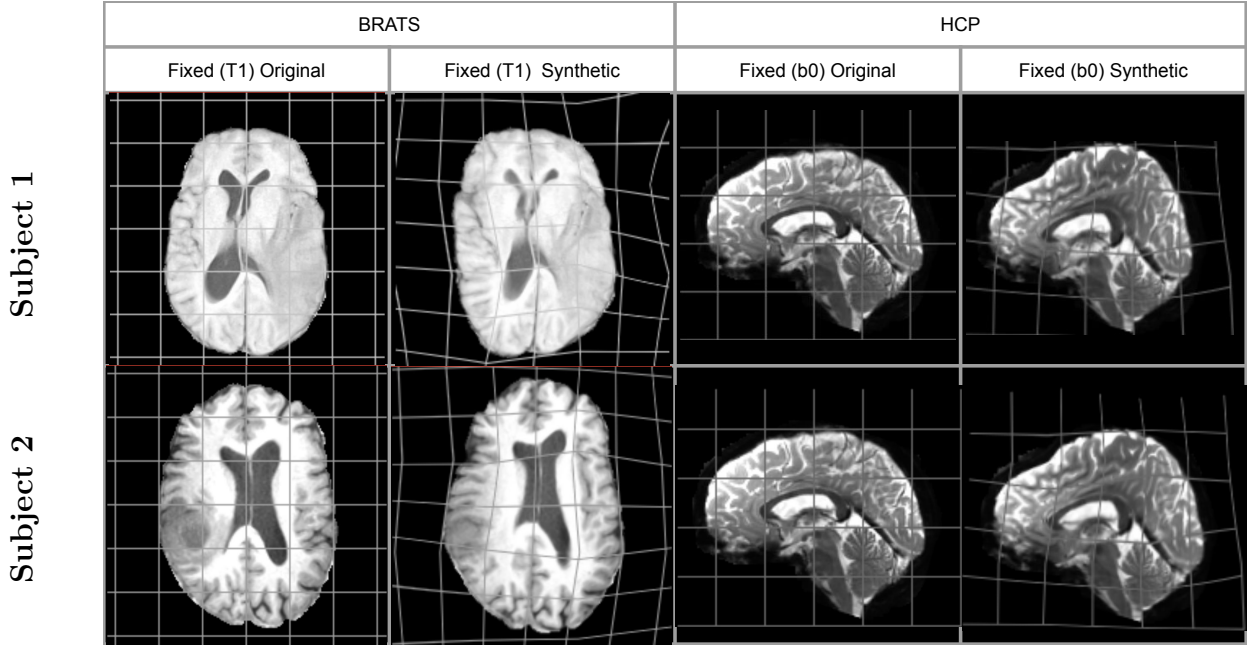
\begin{figure}[ht!] 
    \centering 
    \begin{picture}(0,0) 
    \put(-250,-120){\rotatebox{90}{\textbf{Subject 1}}} 
    \put(-250,-220){\rotatebox{90}{\textbf{Subject 2}}} 
    \end{picture} 
    \includegraphics[page=2,width=\textwidth]{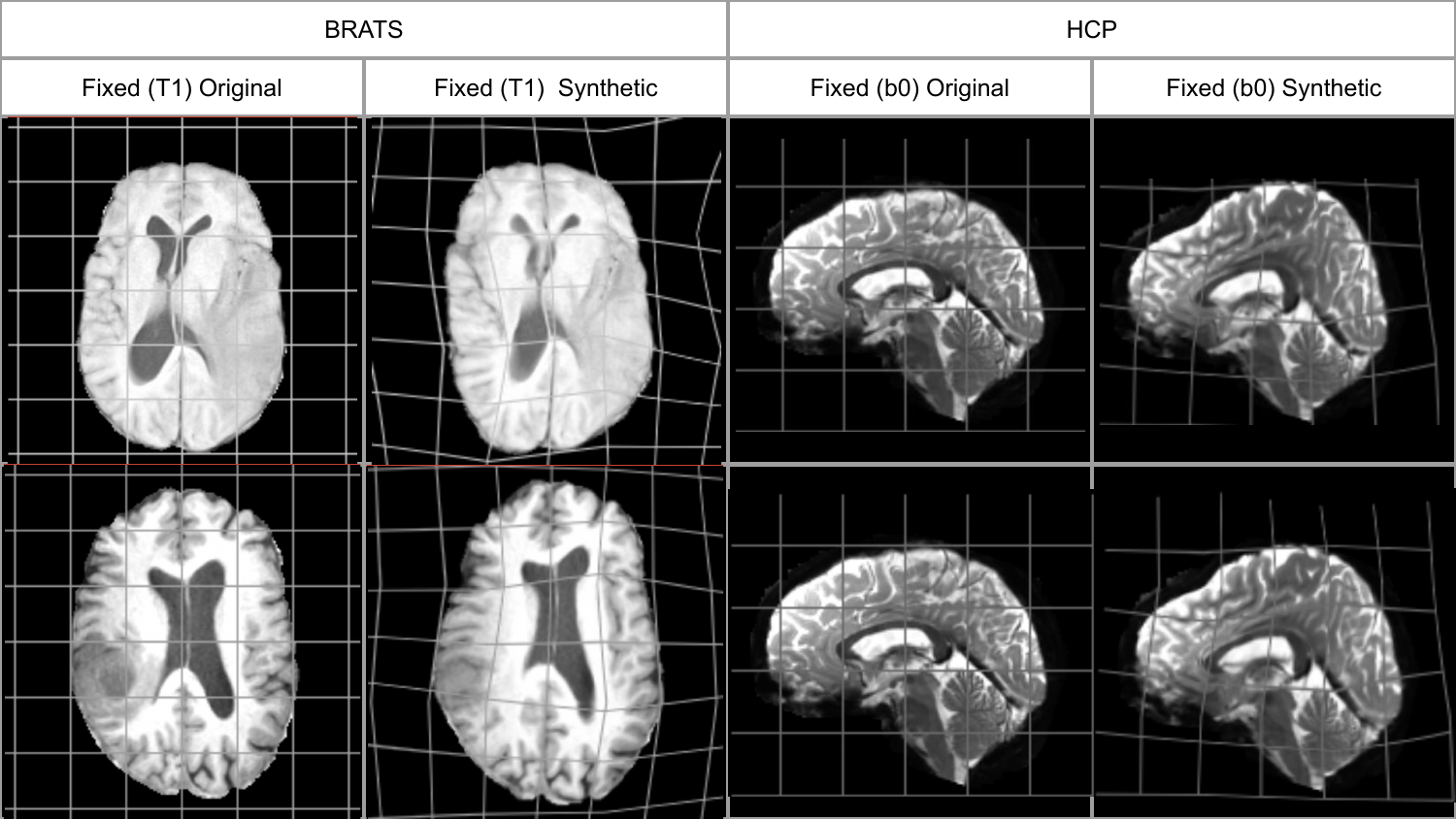} 
    \caption{Synthetic deformation examples on the BRATS and HCP datasets. Each row corresponds to a different patient sample. The grid overlay illustrates the global deformation pattern.} \label{fig:synt_def_example} 
\end{figure}

For datasets containing \underline{pre-registered} pairs (e.g. BRATS and HCP), synthetic transformations are generated to enable a quantitative evaluation of the registration results. For each pair of registered source and target modality images, random B-spline displacement fields are sampled independently using a coarse control grid and interpolated over the full target modality image domain. The control point grid size is fixed to $(4 \times 4 \times 4)$, corresponding to the minimum grid size imposed to support cubic B-splines (degree $(3,3,3)$). Such a coarse grid induces smooth, low-frequency deformations, as each control point influences a large spatial region through the compact support of the spline basis functions, thereby limiting the representation of high-frequency spatial variations. The resulting displacement field is then applied to the target modality image to obtain a synthetically deformed target image. Two samples from the BRATS and HCP datasets are shown in Figure~\ref{fig:synt_def_example}. Control point displacements $\mathbf{c}_{i}$ are resampled independently from a uniform distribution:
\begin{equation}
    \mathbf{c}_{i} \sim \mathcal{U}(-s, s)^3,
\end{equation}
where $s = 15$ voxels (15-19~mm) defines the maximum displacement amplitude along each spatial dimension, thus providing an overall spectrum of small and large patient displacements, corresponding to slipping or intentional motion, representing a range of plausible anatomical intra-subject displacements observed in clinical practice. The dense displacement field is obtained via BSpline interpolation of the control grid over normalized spatial coordinates in $[0,1]^3$, yielding spatially smooth deformation fields without additional explicit regularization. The target modality image (resp. segmentation) is then warped using trilinear (resp. nearest neighbor) interpolation according to the resulting displacement field to create a synthetic target image. Part of the resulting images were checked for anatomical consistency by a biomedical engineer with more than 7 years of experience in medical imaging.

\subsection{Training protocol}
For all the methods that require training, the paradigm applied is the one described in their respective papers and, where applicable, the similarity metric is replaced with the MIND-SSC loss. For consistency and fair comparison, the training is unsupervised for all methods. We accumulate gradients to match an effective batch size of 8 samples and perform validation every 5 training epochs. All models are trained with an Adam optimizer with an initial learning rate of $10^{-4}$ and a learning rate scheduler stepping based on the validation loss with a patience of 10 epochs and decrease factor of 0.5. Early stopping is called when the learning rate falls below $10^{-7}$.
Furthermore, all trainings were performed on the IDRIS Jean-Zay cluster using the pytorch-gpu/py3/2.8.0 environment. 
VoxelMorph, and GradICON trainings were performed on a single NVIDIA Tesla V100 SXM2 with 32~GB VRAM. M2M-GradICON required more memory and used a single NVIDIA A100 SXM4 with 80~GB VRAM.

\subsection{Evaluation Metrics}
\label{sec:evaluation-metrics}
The registration performance is evaluated using overlap-based and boundary-based metrics computed on raw images and anatomical segmentations. As a result, five complementary accuracy metrics are presented hereafter. The Dice similarity coefficient (DSC) is used to measure volumetric overlap between registered and reference segmentations, and is reported as the average across all studied anatomical structures. In the following, it will be noted as $\overline{\mathrm{DSC}}$ for simplicity. Boundary accuracy is assessed using the 95\textsuperscript{th}-percentile Hausdorff Distance (HD95), averaged across all studied anatomical structures and noted as $\overline{\mathrm{HD95}}$ in the following. The value of the MIND-SSC loss is also reported as a surrogate measure of structural similarity to assess the global image improvement. 

The MIND-SSC loss is defined as follows. Given a local neighborhood $\mathcal{N}$, an image $I$, a global noise estimate $\sigma$ and SSD the Sum of Squared Differences, the self-similarity descriptor between two patches $x$ and $y$ in $\mathcal{N}$ is given by:
\begin{equation}
    S(I, x, y) = \mathrm{exp}\left(-\frac{SSD(x, y)}{\sigma^2}\right)
\end{equation}

Given $I$ and $J$ two images defined on an image domain $\Omega$ and $x\in\Omega$ the center of a homologous patch within these two images, the MIND-SSC loss $\mathcal{L}_{MIND-SSC}$ is obtained as:
\begin{equation}
    \mathcal{L}_{MIND-SSC} = \frac{1}{|\mathcal{N}|}\sum_{x\in\Omega}\sum_{y\in\mathcal{N}}|S(I, x, y) - S(J, x, y)|.
\end{equation}

In addition, the determinant of the Jacobian of the displacement field is computed to obtain the folding-ratio (f$\_$ratio) percentage of the proposed transformation. Finally, given that the registration on the BRATS and HCP datasets was computed on synthetically displaced images and that the original displacement field is available, it is interesting to report the relative error of the Root-Mean-Square-Error (DVF rel. error) between the original displacement field and the one predicted for the registration. The result is provided as a percentage value of the difference with respect to the norm of the original displacement field. 

A detailed description of these metrics is given in Table~\ref{tab:evaluation_metrics} where $a$ and $b$ denote binary segmentations; $A_l$ and $B_l$ denote binary segmentation of multilabel segmentations $A$ and $B$ restrained to label $l\in[1; N]$; $d(\cdot,\cdot)$ is the surface distance; $P_{95}$ denotes the 95\textsuperscript{th}-percentile; $\phi$ is the estimated transformation; $J_\phi$ its Jacobian matrix; $\Omega$ the image domain; and $\phi_{\mathrm{pred}}$ and $\phi_{\mathrm{gt}}$ the predicted and ground-truth displacement vector fields (DVFs), respectively.

\begin{table}[htbp] 
\centering 
\caption{Summary of the evaluation metrics.}
\label{tab:evaluation_metrics} 
\small 
\begin{tabularx}{\linewidth}{l X} 
\toprule 
\textbf{Metric} & \textbf{Definition} \\ 
\midrule 
DSC & $\displaystyle \mathrm{DSC}(a,b)= \frac{2|a\cap b|}{|a|+|b|} $ \\
[0.5em] $\overline{\mathrm{DSC}}$ & $\displaystyle \overline{\mathrm{DSC}}(A,B)= \frac{1}{N}\sum_{l=1}^{N}\mathrm{DSC}(A_l, B_l) $ \\
[0.5em] HD95 & $\displaystyle \mathrm{HD95}(a,b)= \max\!\left( P_{95}\!\left(d(a,b)\right), P_{95}\!\left(d(b,a)\right) \right) $ \\
[0.5em] $\overline{\mathrm{HD95}}$ & $\displaystyle \overline{\mathrm{HD95}}(A,B)= \frac{1}{N}\sum_{l=1}^{N}\mathrm{HD95}(A_l, B_l) $ \\
[0.5em] $\mathcal{L}_{MIND-SSC}$ &
$\displaystyle \mathcal{L}_{MIND-SSC} = \frac{1}{|\mathcal{N}|}\sum_{x\in\Omega}\sum_{y\in\mathcal{N}}|S(I, x_i, y) - S(J, x_j, y)|.$ \\
[0.5em] f\_ratio & $\displaystyle 100\cdot \frac{ \#\{x:\det(J_\phi(x))<0\} }{ \#\Omega } $ \\
[0.5em] DVF rel. error & $\displaystyle 100\cdot \frac{ \|\phi_{\mathrm{pred}}-\phi_{\mathrm{gt}}\|_2}{ \|\phi_{\mathrm{gt}}\|_2 } $ \\ 
\bottomrule 
\end{tabularx} 
\end{table}

All metrics are computed by propagating segmentation labels from the moving image using the estimated transformation and comparing them to the reference labels in the target image space. Results are reported as mean and standard deviation among subjects.

In addition, we report computational efficiency metrics such as the runtime, throughput, CPU and GPU usage. Runtime and throughput were measured over 50 repeated inference runs on the same hardware platform. CPU utilization is normalized over the total available system capacity. GPU memory corresponds to peak allocated GPU memory during inference.

All reported metrics were computed using a Linux WSL2 environment (Linux 6.6.87.2) using Python 3.12.3 on a 32-core x86\_64 CPU system equipped with 31.2~GB of RAM and a single NVIDIA GeForce RTX 5090 GPU with 31.8~GB VRAM running CUDA 13.0.

\subsection{Benchmarking Protocol}
All methods are evaluated under a unified benchmarking framework to ensure fair comparison. Identical preprocessing steps, image pairs, and intra-patient registration settings are used across all methods. The same evaluation pipeline is applied uniformly to all outputs.

For each classical registration method, the main set of tunable hyper-parameters was selected and optimized on a subset of 10 cases from the validation set. Hyper-parameter configurations were ranked according to the average Dice similarity coefficient ($\overline{\mathrm{DSC}}$), and the best-performing configuration was subsequently used for all evaluations on the corresponding dataset. The selected hyper-parameters for each method and dataset are reported in Table~\ref{tab:classical_hparams}. As for learning-based methods the large training cost makes it intractable to perform extensive hyper-parameter grid-search,  we thus decided to evaluate them using their default hyper-parameters, as proposed in their respective articles. Unlike the other learning-based methods, the pretrained multiGradICON model imposes a fixed input size of $175\times175\times175$ voxels. Consequently, all images evaluated with this model were cropped to satisfy this constraint while conserving the maximum amount of segmented organs within the bounds of the test images. The resulting performance should therefore be interpreted in light of this architectural limitation.

\begin{table*}[t] 
\centering 
\caption{Best hyper-parameter configurations selected for each classical registration method and dataset.} 
\label{tab:classical_hparams} 
\small 
\begin{tabularx}{\linewidth}{ 
>{\raggedright\arraybackslash}p{2.2cm} 
>{\raggedright\arraybackslash}X c c c } 
\toprule 
Method & Hyper-parameter & BRATS & HCP & IMAG2 (pelvis) \\ 
\midrule 
\multirow[c]{4}{2.2cm}{\centering ANTs SyN} 
& grad\_step & 0.2 & 0.2 & 0.1 \\ 
& flow\_sigma & 3.0 & 2.0 & 3.0 \\ 
& total\_sigma & 0.5 & 0.0 & 0.0 \\ 
& reg\_iterations & (80,40,20) & (80,40,20) & (80,40,20) \\ 
\midrule 
\multirow[c]{4}{2.2cm}{\centering Elastix} 
& maximum\_number\_of\_iterations & 512 & 256 & 512 \\ 
& number\_of\_resolutions & 3 & 3 & 4 \\ 
& number\_of\_spatial\_samples & 2048 & 2048 & 2048 \\ 
& bspline\_final\_grid\_spacing (mm) & 32.0 & 16.0 & 32.0 \\ 
\midrule 
\multirow[c]{3}{2.2cm}{\centering Log-Demons} 
& n\_iter & 30 & 80 & 30 \\ 
& sigma\_update & 1.5 & 1.5 & 1.5 \\ 
& sigma\_field & 1.0 & 1.0 & 1.0 \\ 
\bottomrule 
\end{tabularx} 
\end{table*}

\section{Results}
\label{sec:Results}

This section presents the quantitative and qualitative results of the benchmark described in Section~\ref{sec:Methods}. The quantitative section focuses on the analyses of the metrics reported in Section~\ref{sec:evaluation-metrics} applied to all the registration methods, including their anatomy-based versions described in paragraph ``Anatomically guided registration'' of Section~\ref{par:anatomically-guided-registration}. Qualitative results are then presented in Section \ref{sec:qualitative}.

\subsection{Registration Accuracy} 
This section presents the quantitative results for all registration approaches defined in Section~\ref{sec:evaluated-registration-methods}. Given the difference in input data between the anatomy-guided and image-guided registration experiments, the presentation of the respective results is divided in Sections~\ref{sec:image-guided-registration-results} and~\ref{sec:anatomy-guided-registration-results}

\begin{table*}[ht]
\centering
\caption{Registration accuracy results on BRATS.}
\label{tab:results_accuracy_brats}

\scriptsize
\setlength{\tabcolsep}{2pt}
\renewcommand{\arraystretch}{1.15}

\resizebox{\textwidth}{!}{%
\begin{tabular}{lccccc}
\toprule

Method
& $\overline{\mathrm{DSC}}$ (\%) $\uparrow$
& $\overline{\mathrm{HD95}}$ (mm) $\downarrow$
& $\mathcal{L}_{MIND-SSC}$ $\downarrow$
& DVF rel. error (\%)$\downarrow$
& f\_ratio $\cdot 10^{2}$ (\%) $\downarrow$ \\

\midrule
Unregistered
& 42.6 $\pm$ 13.5 & 7.6 $\pm$ 2.1 & 47.6 $\pm$ 5.5 & 100 & -- \\

\midrule

ANTs SyN
& 59.2 $\pm$ 19.0 & 7.1 $\pm$ 17.8 & 50.8 $\pm$ 15.8 & 94.3 $\pm$ 2.3 & 0.0003 $\pm$ 0.0012 \\

Elastix
& 74.8 $\pm$ 9.6 & 2.4 $\pm$ 1.0 & 44.9 $\pm$ 9.5 & 99.0 $\pm$ 14.3 & \textbf{\underline{0.0000 $\pm$ 0.0000}} \\

Demons
& 41.8 $\pm$ 13.3 & 7.7 $\pm$ 2.1 & 38.5 $\pm$ 5.2 & 99.4 $\pm$ 0.3 & 0.0001 $\pm$ 0.0004 \\

\midrule

VoxelMorph
& 62.3 $\pm$ 11.1 & 3.9 $\pm$ 1.4 & \textbf{\underline{10.8 $\pm$ 2.4}} & 88.5 $\pm$ 2.5 & 2.5562 $\pm$ 0.4611 \\

GradICON
& 77.0 $\pm$ 8.0 & 1.9 $\pm$ 0.5 & 12.2 $\pm$ 3.2 & 81.3 $\pm$ 4.7 & 3.2633 $\pm$ 0.7565 \\

M2M\_GradICON
& 55.1 $\pm$ 12.5 & 4.9 $\pm$ 1.7 & 23.0 $\pm$ 3.8 & 89.5 $\pm$ 2.4 & 0.0018 $\pm$ 0.0046 \\

multiGradICON
& 64.0 $\pm$ 11.9 & 3.9 $\pm$ 1.8 & 20.7 $\pm$ 4.8 & \textbf{\underline{81.2 $\pm$ 6.0}} & 0.0061 $\pm$ 0.0063 \\

\midrule

ANTs SyN seg.
& \textbf{\underline{93.3 $\pm$ 3.9}} & \textbf{\underline{1.1 $\pm$ 0.3}} & 87.3 $\pm$ 12.0 & 99.7 $\pm$ 0.3 & 0.0000 $\pm$ 0.0000 \\

GradICON seg.
& 76.7 $\pm$ 9.9 & 2.0 $\pm$ 0.7 & 16.5 $\pm$ 3.6 & 82.0 $\pm$ 4.6 & 0.0268 $\pm$ 0.0321 \\

\bottomrule
\end{tabular}
}
\end{table*}

\begin{table*}[ht]
\centering
\caption{Registration accuracy results on HCP.}
\label{tab:results_accuracy_hcp}

\scriptsize
\setlength{\tabcolsep}{2pt}
\renewcommand{\arraystretch}{1.15}

\resizebox{\textwidth}{!}{%
\begin{tabular}{lccccc}
\toprule

Method
& $\overline{\mathrm{DSC}}$ (\%) $\uparrow$
& $\overline{\mathrm{HD95}}$ (mm) $\downarrow$
& $\mathcal{L}_{MIND-SSC}$ $\downarrow$
& DVF rel. error (\%)$\downarrow$
& f\_ratio $\cdot 10^{2}$ (\%) $\downarrow$ \\

\midrule
Unregistered
& 19.7 $\pm$ 4.6 & 9.5 $\pm$ 1.5 & 116.2 $\pm$ 9.3 & 100 & -- \\

\midrule

ANTs SyN
& 17.0 $\pm$ 11.6 & 15.4 $\pm$ 4.8 & 120.1 $\pm$ 9.0 & 99.5 $\pm$ 0.5 & 0.0135 $\pm$ 0.0135 \\

Elastix
& 44.7 $\pm$ 9.8 & 5.9 $\pm$ 2.2 & 121.6 $\pm$ 7.6 & 211.3 $\pm$ 38.2 & 0.0169 $\pm$ 0.0707 \\

Demons
& 29.7 $\pm$ 8.2 & 8.4 $\pm$ 1.7 & 115.2 $\pm$ 9.5 & 98.6 $\pm$ 0.5 & 0.0017 $\pm$ 0.0022 \\

\midrule

VoxelMorph
& 54.7 $\pm$ 3.0 & 3.0 $\pm$ 0.3 & \textbf{\underline{8.7 $\pm$ 1.4}} & 142.3 $\pm$ 8.3 & 6.4165 $\pm$ 0.6430 \\

GradICON
& 57.0 $\pm$ 4.4 & 3.7 $\pm$ 0.5 & 17.7 $\pm$ 2.6 & 211.2 $\pm$ 14.3 & 4.9456 $\pm$ 0.2164 \\

M2M\_GradICON
& 45.0 $\pm$ 6.6 & 4.4 $\pm$ 4.2 & 91.8 $\pm$ 7.4 & \textbf{\underline{94.0 $\pm$ 3.1}} & 0.0011 $\pm$ 0.0042 \\

multiGradICON
& 22.3 $\pm$ 7.5 & 10.4 $\pm$ 2.1 & 67.4 $\pm$ 3.9 & 155.9 $\pm$ 14.8 & 0.6305 $\pm$ 0.2771 \\

\midrule

ANTs SyN seg.
& \textbf{\underline{86.3 $\pm$ 3.6}} & 1.8 $\pm$ 0.5 & 139.1 $\pm$ 11.2 & 98.5 $\pm$ 0.8 & \textbf{\underline{0.0001 $\pm$ 0.0005}} \\

GradICON seg.
& 74.6 $\pm$ 3.3 & \textbf{\underline{1.6 $\pm$ 0.3}} & 21.8 $\pm$ 2.0 & 162.2 $\pm$ 10.9 & 2.3012 $\pm$ 0.1990 \\

\bottomrule
\end{tabular}
}
\end{table*}

\begin{table*}[ht]
\centering
\caption{Registration accuracy results on IMAG2 (pelvis).}
\label{tab:results_accuracy_pelvis}

\scriptsize
\setlength{\tabcolsep}{2pt}
\renewcommand{\arraystretch}{1.15}

\resizebox{\textwidth}{!}{%
\begin{tabular}{lccccc}
\toprule

Method
& $\overline{\mathrm{DSC}}$ (\%) $\uparrow$
& $\overline{\mathrm{HD95}}$ (mm) $\downarrow$
& $\mathcal{L}_{MIND-SSC}$ $\downarrow$
& DVF rel. error (\%)$\downarrow$
& f\_ratio $\cdot 10^{2}$ (\%) $\downarrow$ \\

\midrule
Unregistered
& 67.0 $\pm$ 14.3 & 8.0 $\pm$ 4.2 & 17.3 $\pm$ 9.7 & -- & -- \\

\midrule

ANTs SyN
& 66.2 $\pm$ 12.8 & 7.9 $\pm$ 3.7 & 42.2 $\pm$ 21.3 & -- & 0.0213 $\pm$ 0.0846 \\

Elastix
& 59.2 $\pm$ 13.8 & 10.3 $\pm$ 5.2 & 26.9 $\pm$ 17.6 & -- & 0.0134 $\pm$ 0.0536 \\

Demons
& 61.3 $\pm$ 14.1 & 9.0 $\pm$ 4.1 & 23.8 $\pm$ 11.9 & -- & 0.0912 $\pm$ 0.0781 \\

\midrule

VoxelMorph
& 65.1 $\pm$ 14.8 & 7.9 $\pm$ 3.7 & \textbf{\underline{14.6 $\pm$ 12.1}} & -- & 1.8370 $\pm$ 1.0131 \\

GradICON
& 67.1 $\pm$ 14.3 & 8.0 $\pm$ 4.2 & 17.2 $\pm$ 9.7 & -- & \textbf{\underline{0.0000 $\pm$ 0.0000}} \\

M2M\_GradICON
& 61.6 $\pm$ 11.7 & 9.0 $\pm$ 4.2 & 23.4 $\pm$ 12.4 & -- & 0.0000 $\pm$ 0.0000 \\

multiGradICON
& 60.8 $\pm$ 10.4 & 8.7 $\pm$ 3.9 & 81.0 $\pm$ 31.2 & -- & 0.1887 $\pm$ 0.1674 \\

\midrule

ANTs SyN seg.
& \textbf{\underline{81.2 $\pm$ 9.5}} & \textbf{\underline{5.5 $\pm$ 4.4}} & 40.0 $\pm$ 18.2 & -- & 0.0025 $\pm$ 0.0087 \\

GradICON seg.
& 67.1 $\pm$ 13.9 & 7.9 $\pm$ 4.0 & 19.4 $\pm$ 10.1 & -- & 0.0010 $\pm$ 0.0023 \\

\bottomrule
\end{tabular}
}
\end{table*}

\subsubsection{Image-guided registration}
\label{sec:image-guided-registration-results}
Tables~\ref{tab:results_accuracy_brats}--\ref{tab:results_accuracy_pelvis} summarize the registration performance of all evaluated methods using the metrics defined in Section~\ref{sec:evaluation-metrics}. Higher $\overline{\mathrm{DSC}}$ values indicate better overlap, whereas lower $\overline{\mathrm{HD95}}$, $\mathcal{L}_{MIND-SSC}$, DVF error, and $f_{\mathrm{ratio}}$ values indicate superior performance. Note that DVF error is only available for BRATS and HCP because these datasets were generated using known synthetic deformations. The IMAG2 (pelvis) dataset corresponds to a real registration task and therefore does not provide a deformation ground truth. Across all datasets, substantial variability is observed between methods and anatomical regions, highlighting the difficulty of multimodal deformable registration without explicit anatomical supervision. Moreover, the different metrics reveal important trade-offs between segmentation overlap, image-based similarity, deformation recovery, and transformation regularity. On the BRATS dataset, all methods except Demons achieved substantial improvements over the initial misalignment. Among the classical approaches, Elastix obtained the strongest overlap-based performance, reaching a $\overline{\mathrm{DSC}}$ of $74.8 \pm 9.6$ and an $\overline{\mathrm{HD95}}$ of $2.4 \pm 1.0$ mm. However, its DVF error remained high ($99.0 \pm 14.3$), indicating poor recovery of the underlying synthetic deformation despite good anatomical alignment. Demons achieved the lowest MIND-SSC loss among classical methods ($38.5 \pm 5.2$) but failed to improve $\overline{\mathrm{DSC}}$ and $\overline{\mathrm{HD95}}$ compared with the unregistered baseline. Among learning-based approaches, GradICON produced the best geometric registration accuracy, achieving the highest $\overline{\mathrm{DSC}}$ ($77.0 \pm 8.0$) and the lowest $\overline{\mathrm{HD95}}$ ($1.9 \pm 0.5$ mm). VoxelMorph achieved the lowest MIND-SSC loss overall ($10.8 \pm 2.4$), suggesting excellent local structural alignment, while multiGradICON obtained the lowest DVF error ($81.2 \pm 6.0$), closely followed by GradICON ($81.3 \pm 4.7$). In terms of topology preservation, the classical methods produced almost perfectly diffeomorphic transformations with negligible folding ratios, whereas VoxelMorph and GradICON exhibited substantial folding ($f_{\mathrm{ratio}}=2.56$ and $3.26$, respectively). M2M-GradICON represented a compromise between both families, achieving low folding ($0.0018$) at the cost of reduced registration accuracy. 

The HCP dataset exhibits a similar pattern but with a greater separation between methods. ANTs SyN degraded alignment relative to the baseline, yielding lower $\overline{\mathrm{DSC}}$ ($17.0 \pm 11.6$), higher $\overline{\mathrm{HD95}}$ ($15.4 \pm 4.8$ mm), and increased MIND-SSC loss ($120.1 \pm 9.0$). Elastix improved overlap considerably ($\overline{\mathrm{DSC}}$ $44.7 \pm 9.8$), but again showed very poor deformation recovery with a DVF error exceeding $200$. Among the learning-based methods, GradICON achieved the highest $\overline{\mathrm{DSC}}$ ($57.0 \pm 4.4$), whereas VoxelMorph obtained the lowest $\overline{\mathrm{HD95}}$ ($3.0 \pm 0.3$ mm) and lowest MIND-SSC loss ($8.7 \pm 1.4$). By contrast, M2M-GradICON produced the most accurate recovery of the synthetic deformation, obtaining the lowest DVF error ($94.0 \pm 3.1$), substantially outperforming all competing approaches. MultiGradICON showed limited improvements over the baseline, performing poorly across $\overline{\mathrm{DSC}}$, $\overline{\mathrm{HD95}}$, and MIND-SSC loss despite maintaining a relatively low folding ratio. 

The IMAG2 (pelvis) dataset proved considerably more challenging than the synthetic brain benchmarks. Most methods produced only marginal improvements relative to the unregistered baseline, and several degraded performance. Among the classical methods, ANTs SyN preserved the baseline overlap ($\overline{\mathrm{DSC}}$ $66.2 \pm 12.8$) and $\overline{\mathrm{HD95}}$ ($7.9 \pm 3.7$ mm) but substantially increased the MIND-SSC loss ($42.2 \pm 21.3$), suggesting poorer multimodal structural consistency. Elastix and Demons generally reduced overlap performance relative to the baseline. Among learning-based methods, GradICON achieved the highest $\overline{\mathrm{DSC}}$ ($67.1 \pm 14.3$), matching the baseline performance while maintaining a low MIND-SSC loss ($17.2 \pm 9.7$). VoxelMorph produced the lowest MIND-SSC loss value overall ($14.6 \pm 12.1$) and competitive $\overline{\mathrm{HD95}}$ ($7.9 \pm 3.7$ mm), but at the expense of a large folding ratio ($1.84 \pm 1.01$). By contrast, GradICON and M2M-GradICON generated essentially fold-free transformations ($f_{\mathrm{ratio}}\approx0$) while maintaining comparable registration accuracy. MultiGradICON performed poorly on this dataset, yielding a substantially increased MIND-SSC loss ($81.0 \pm 31.2$) and lower overlap than the baseline. 

Overall, the results demonstrate that registration performance depends strongly on the anatomical region and imaging characteristics. On the synthetic BRATS and HCP datasets, learning-based approaches outperformed classical methods in terms of overlap accuracy and image consistency. However, a distinction emerges between methods that best recover the known synthetic deformation and those that optimize image-based registration metrics. In particular, M2M-GradICON and multiGradICON generally achieved the lowest DVF errors and folding ratios, whereas GradICON and VoxelMorph achieved superior $\overline{\mathrm{DSC}}$, $\overline{\mathrm{HD95}}$, and MIND-SSC loss scores. This suggests that accurate reconstruction of the synthetic deformation does not necessarily translate into the best anatomical alignment under the chosen evaluation criteria. On the real IMAG2 (pelvis) dataset, where no deformation ground truth is available, all methods exhibited more modest gains, indicating that the registration task is substantially more difficult. 

Among the unsupervised methods, GradICON provided the most consistent behavior across datasets, combining strong overlap accuracy with low MIND-SSC loss, while maintaining near-diffeomorphic transformations on IMAG2 (pelvis).

\subsubsection{Anatomy-guided registration} 
\label{sec:anatomy-guided-registration-results}
The lower part of Tables~\ref{tab:results_accuracy_brats}--\ref{tab:results_accuracy_pelvis} reports the performance of registration methods when anatomical segmentations are explicitly incorporated into the optimization process. Because these approaches exploit supervision unavailable to the unsupervised methods, they should be interpreted separately and may be viewed as an approximate upper bound on the achievable registration accuracy when anatomical annotations are available.

On the BRATS dataset, ANTs SyN seg.\ achieved the strongest overlap-based performance, obtaining a $\overline{\mathrm{DSC}}$ of $93.3 \pm 3.9$ and the lowest $\overline{\mathrm{HD95}}$ of $1.1 \pm 0.3$ mm. However, these improvements were accompanied by a very large MIND-SSC loss ($87.3 \pm 12.0$), substantially higher than all learning-based methods and even higher than the unregistered baseline. Furthermore, despite the excellent overlap performance, the DVF error remained high ($99.7 \pm 0.3$), indicating poor recovery of the underlying synthetic deformation. By contrast, GradICON seg.\ achieved lower overlap ($\overline{\mathrm{DSC}}$ $76.7 \pm 9.9$) but substantially improved image-based consistency, yielding a much lower MIND-SSC loss value ($16.5 \pm 3.6$) than ANTs SyN seg.\ while remaining higher than the standard GradICON version. Its DVF error ($82.0 \pm 4.6$) was also markedly lower than that of ANTs SyN seg., indicating a deformation closer to the ground-truth transformation. Topology preservation further distinguished the two approaches: ANTs SyN seg.\ remained essentially fold-free ($f_{\mathrm{ratio}} \approx 0$), whereas GradICON seg.\ introduced a small amount of folding ($0.0268 \pm 0.0321$). These results illustrate that maximizing anatomical overlap does not necessarily imply improved multimodal consistency or more accurate recovery of the true deformation. 

The same trend is observed on HCP. ANTs SyN seg.\ achieved the highest $\overline{\mathrm{DSC}}$ ($86.3 \pm 3.6$) and near-optimal topology preservation with an almost-zero folding ratio ($0.0001 \pm 0.0005$). However, it also produced the highest MIND-SSC loss ($139.1 \pm 11.2$) among all evaluated methods and a DVF error of $98.5 \pm 0.8$. GradICON seg.\ achieved slightly lower overlap ($\overline{\mathrm{DSC}}$ $74.6 \pm 3.3$) but produced the best $\overline{\mathrm{HD95}}$ ($1.6 \pm 0.3$ mm), indicating more accurate boundary alignment. More importantly, its MIND-SSC loss ($21.8 \pm 2.0$) was dramatically lower than that obtained by ANTs SyN seg., suggesting substantially better preservation of multimodal structural information. Nevertheless, its DVF error remained high ($162.2 \pm 10.9$), indicating that neither anatomy-guided approach accurately recovered the known synthetic deformation despite achieving strong registration metrics. This is probably because, while the anatomical input provides a solid basis for the registration of the segmentations, no constraint is available outside of the segmented area. As such, any field transforming the partial field (on the segmented areas) into a near-diffeomorphic one is considered a correct solution, thus limiting their ability to recover the actual deformation.Compared with the unsupervised variants, anatomy-guided optimization primarily improved geometric alignment measures while providing less clear benefits in terms of deformation recovery. 

On the IMAG2 (pelvis) dataset, where no deformation ground truth is available, ANTs SyN seg.\ again produced the highest overlap accuracy, achieving a $\overline{\mathrm{DSC}}$ of $81.2 \pm 9.5$ and the lowest $\overline{\mathrm{HD95}}$ ($5.5 \pm 4.4$ mm). However, this gain was accompanied by a relatively large MIND-SSC loss ($40.0 \pm 18.2$), suggesting that the optimization focused strongly on aligning annotated structures. GradICON seg.\ achieved substantially lower overlap, matching the baseline but preserved multimodal consistency more effectively, obtaining a considerably lower MIND-SSC loss value ($19.4 \pm 10.1$). Both methods produced nearly diffeomorphic transformations, with very small folding ratios, although GradICON seg.\ generated the smoother deformation field overall ($0.0010 \pm 0.0023$ versus $0.0025 \pm 0.0087$). As observed on the synthetic datasets, the improvements obtained through segmentation guidance were much more pronounced on overlap-based metrics than on image-based similarity measures. 

Overall, anatomy-guided registration substantially improves geometric alignment metrics, particularly $\overline{\mathrm{DSC}}$ and $\overline{\mathrm{HD95}}$, and enables very accurate matching of annotated anatomical structures. However, the results consistently reveal a trade-off between segmentation overlap and multimodal image consistency. ANTs SyN seg.\ systematically achieves the highest overlap scores across all datasets while maintaining nearly fold-free transformations, but often produces substantially worse MIND-SSC loss values and does not consistently recover the underlying synthetic deformation. Conversely, GradICON seg.\ generally yields lower overlap performance but produces markedly lower MIND-SSC loss and, on BRATS, substantially better DVF recovery. These findings suggest that optimizing directly for anatomical correspondence can bias the registration towards the labeled structures and may not necessarily yield globally consistent multimodal alignments. Consequently, anatomy-guided methods should be evaluated using multiple complementary metrics rather than overlap measures alone.

\subsection{Qualitative results}\label{sec:qualitative}
\begin{figure}[ht!]
    \centering
    \includegraphics[page=3, width=\textwidth]{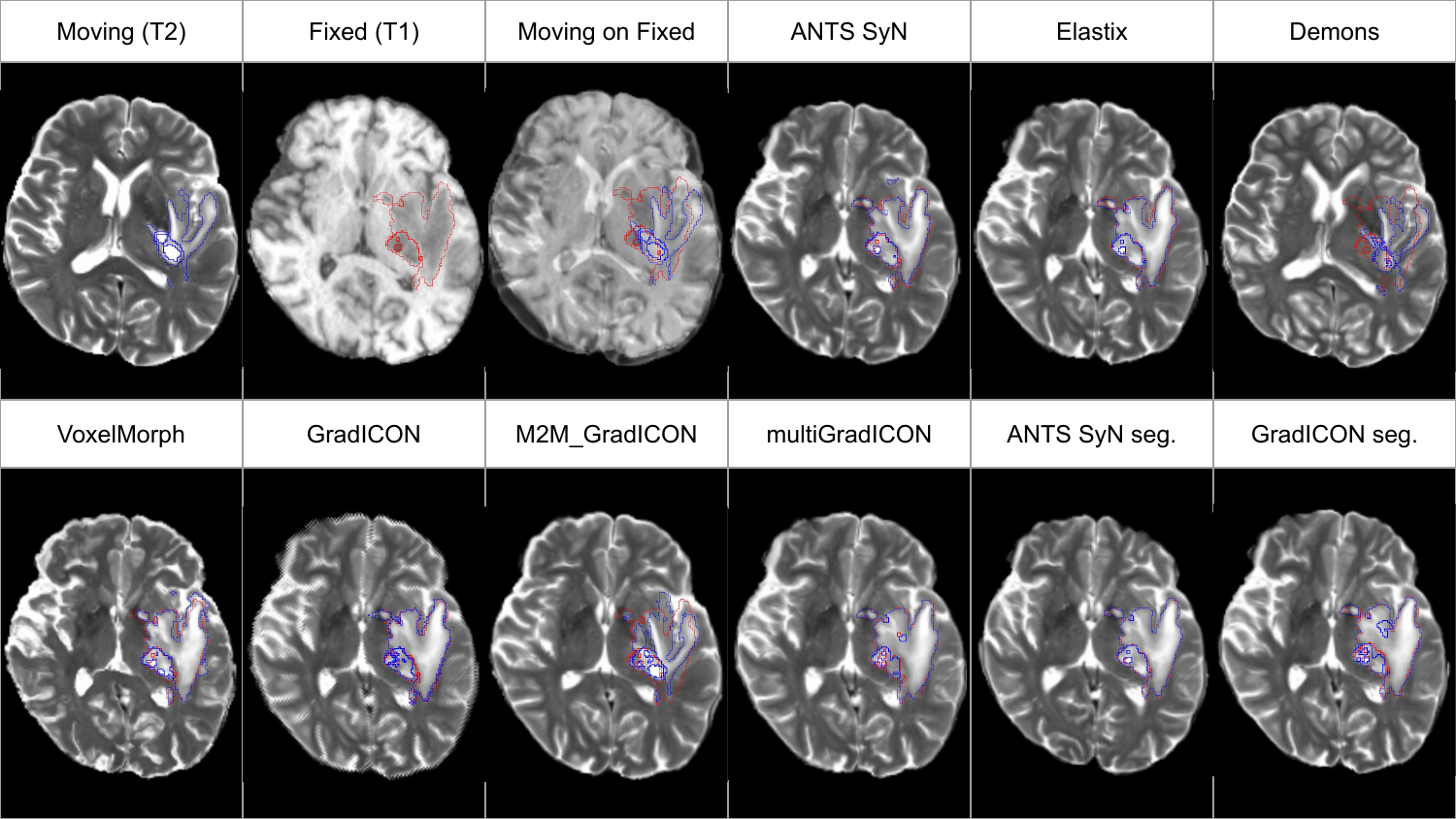}
    \caption{Qualitative registration result of case BraTS-GLI-01259-000 (BRATS) with segmentation masks of the whole tumor. \textbf{Top row}: Moving T2 image (blue segmentation), Fixed T1 image (red segmentation), Initial overlay, overlay of the fixed and registerd image for classic approaches. \textbf{Bottom row}: Overlay of fixed and registered images from deep learning approaches as well as the comparison of GradICON and Ants SyN with additional anatomical guidance.}
    \label{fig:qualitative_results_BRATS}
\end{figure}

\begin{figure}[h!]
    \centering
    \includegraphics[width=\textwidth]{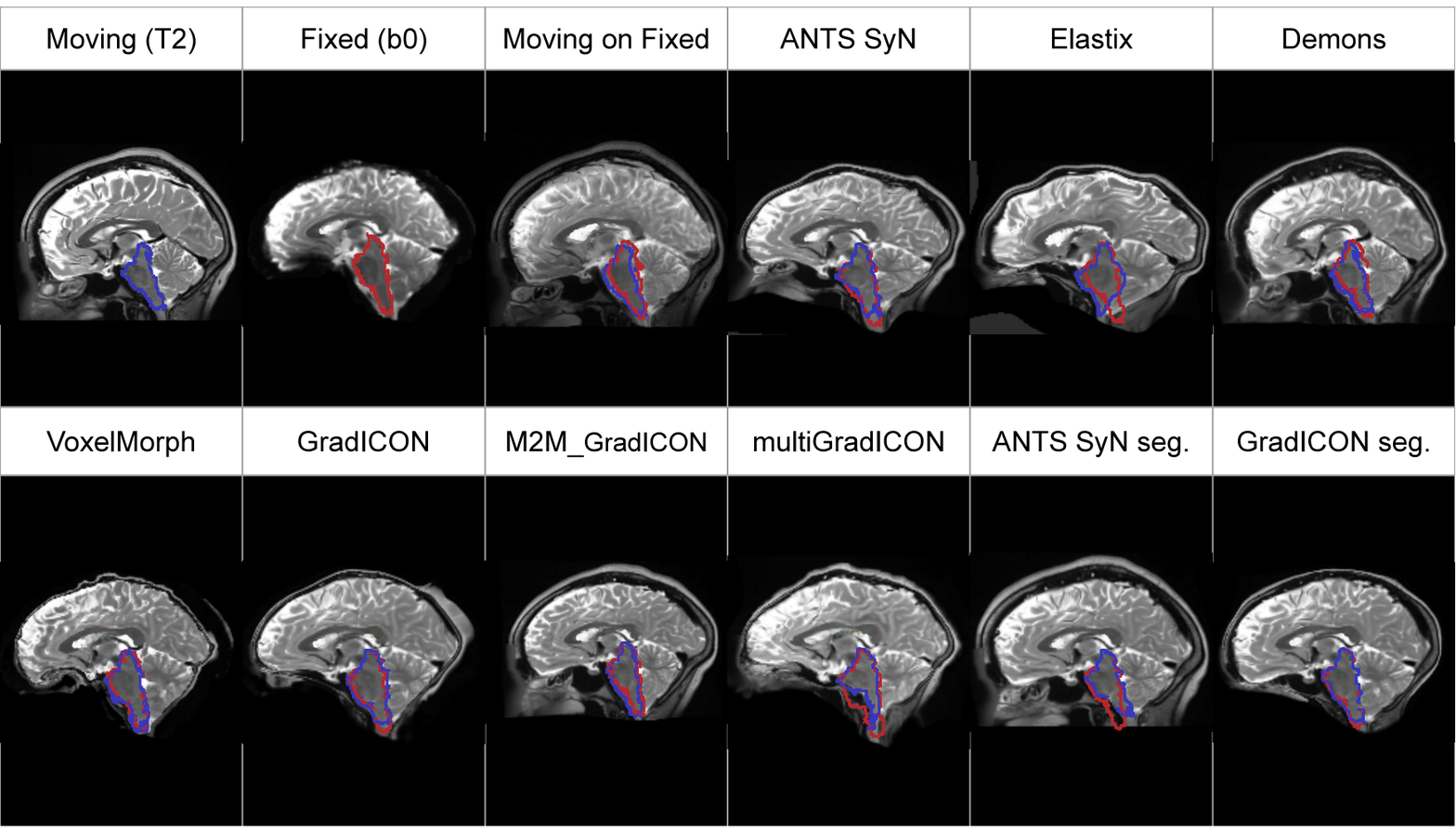}
    \caption{Qualitative registration result of case HCP-792564 (HCP) with segmentation masks of the brain stem. \textbf{Top row}: Moving T2 image (blue segmentation), Fixed T1 image (red segmentation), Initial overlay, overlay of the fixed and registerd image for classic approaches. \textbf{Bottom row}: Overlay of fixed and registered images from deep learning approaches as well as the comparison of GradICON and Ants SyN with additional anatomical guidance.}
    \label{fig:qualitative_results_HCP}
\end{figure}

\begin{figure}[h!]
    \centering
    \includegraphics[page=9, width=\textwidth]{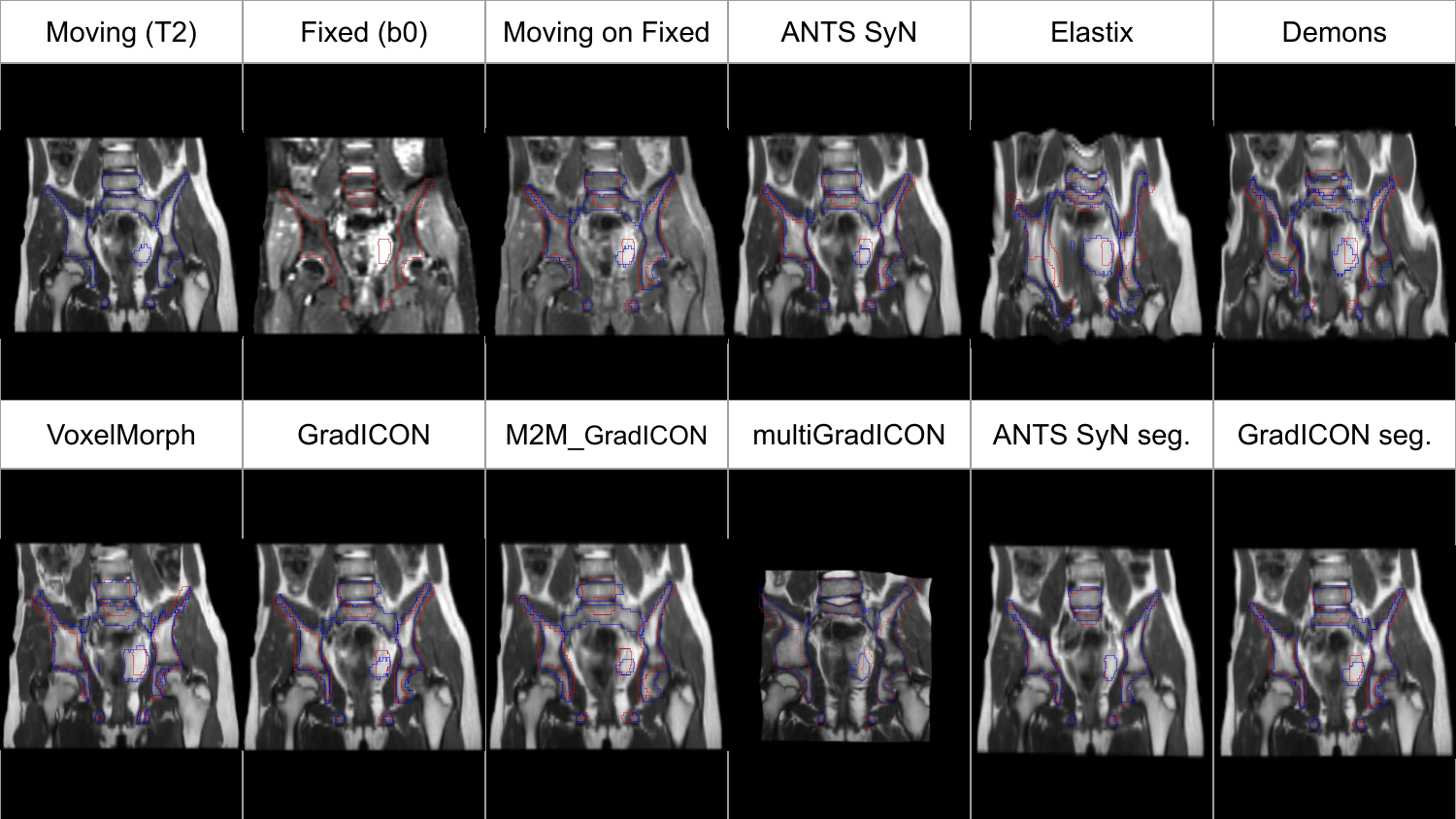}
    \caption{Qualitative registration result of case 1-2-42\_190415 of the IMAG2 (pelvis) dataset. Segmentations include the Pelvis, L5 vertebra, Sacrum and bladder. \textbf{Top row}: Moving T2 image (blue segmentation), Fixed T1 image (red segmentation), Initial overlay, overlay of the fixed and registered image for classic approaches. \textbf{Bottom row}: Overlay of fixed and registered images from deep learning approaches as well as the comparison of ANTS SyN seg. \ and GradICON seg. The visible reduced size of the multiGradICON image is caused by the cropping of the images to fit multiGradICON's input size parameters.}
    \label{fig:qualitative_results_Pelvis}
\end{figure}

Figures~\ref{fig:qualitative_results_BRATS},~\ref{fig:qualitative_results_HCP} and \ref{fig:qualitative_results_Pelvis} show the qualitative registration results for the three employed datasets, compared across all the methods described in Section~\ref{sec:evaluated-registration-methods}. The subjects selected for display are the ones showing --- regardless of the registration method --- the highest improvement (largest difference between pre- and post-registration) on global Dice score for each dataset.

Figure~\ref{fig:qualitative_results_BRATS} shows improved registration results for ANTs SyN and Elastix. The largest structure is globally matched and the smallest inner one partly overlaps. The result of the Demons registration remains questionable. As for the deep-learning based algorithms, similar results are obtained. General overlap of the brain as a whole and the largest structure is obtained for all methods with the exception of M2M\_GradICON which has more limited success as summarized in the quantitative table. Considering ANTs SyN Seg., the structural overlap of all the visible object is striking and has the overall smallest differences.

Figure~\ref{fig:qualitative_results_HCP} reports the registration results on the HCP dataset focusing on the brain stem. The displayed patient mostly shows limited improvement with unrealistic registration of the skull of the T2 image and a poor overlap of the other structures in the brain. The ANTs SyN Seg. and GradICON Seg. showcase the best improvement in the structural overlap even though a significant misalignment subsists on the brain stem on ANTs SyN seg.

Figure~\ref{fig:qualitative_results_Pelvis} presents the results on a case from the IMAG2 (pelvis) dataset. At first glance, it is clear that the misalignment between the structues is less explicit than for the other datasets. As such, most of the registrations do not seem to significantly improve the results and some like Elastix, Demons and multiGradICON deteriorate the alignment of the pelvis. ANTs SyN seg. is the only method that improves the alignment of all the labeled structures.

\subsection{Computational Performance}
Table~\ref{tab:registration_benchmark} reports computational efficiency on the HCP dataset. Learning-based methods were substantially faster than iterative optimization-based registration algorithms. VoxelMorph achieved the lowest runtime at $0.028$~s per registration pair and the highest throughput of $35.5$ registrations/s. multiGradICON also demonstrated high efficiency with $0.179$~s runtime and $5.59$~registrations/s.

By contrast, classical methods required substantially longer execution times. Elastix was the slowest approach with $22.7$~s runtime, followed by ANTs SyN at $8.78$~s. Demons achieved moderate runtime performance ($4.38$~s) but exhibited the highest CPU utilization ($90.8\%$). Learning-based methods relied on GPU acceleration and required between $7.60$ GB and $24.56$ GB of GPU memory, whereas classical methods operated exclusively on CPU resources.

Overall, the results highlight a clear trade-off between registration accuracy and computational efficiency. While segmentation-guided classical approaches achieved the highest accuracy, learning-based methods enabled inference times that were two to three orders of magnitude faster.

\begin{table}[t]
\centering
\caption{Computational performance comparison of registration methods on the HCP dataset.}
\label{tab:registration_benchmark}

\renewcommand{\arraystretch}{1.1}
\setlength{\tabcolsep}{4pt}

\begin{tabular}{lcccc}
\toprule

\textbf{Method} &
\textbf{Runtime} $\downarrow$ &
\textbf{Throughput} $\uparrow$ &
\textbf{CPU} $\downarrow$ &
\textbf{GPU} $\downarrow$ \\

&
(s) &
(reg/s) &
(\%) &
(GB) \\

\midrule

ANTs SyN
& 8.78 $\pm$ 0.29
& 0.114 $\pm$ 0.004
& 24.54 $\pm$ 0.81
& 0.00 $\pm$ 0.00
\\

Elastix
& 22.7 $\pm$ 0.42
& 0.044 $\pm$ 0.000
& 45.5 $\pm$ 0.4
& 0.00 $\pm$ 0.00
\\

Demons
& 4.38 $\pm$ 0.08
& 0.228 $\pm$ 0.004
& 90.8 $\pm$ 1.00
& 0.00 $\pm$ 0.00
\\

VoxelMorph
& \textbf{0.028 $\pm$ 0.001}
& \textbf{35.5 $\pm$ 0.7}
& 3.20 $\pm$ 0.40
& \textbf{7.60 $\pm$ 0.00}
\\

GradICON
& 0.350 $\pm$ 0.019
& 2.86 $\pm$ 0.15
& \textbf{3.16 $\pm$ 0.07}
& 9.68 $\pm$ 0.00
\\

M2M-GradICON
& 0.713 $\pm$ 0.010
& 1.40 $\pm$ 0.02
& \textbf{3.18 $\pm$ 0.02}
& 24.56 $\pm$ 0.00
\\

multiGradICON
& 0.179 $\pm$ 0.003
& 5.59 $\pm$ 0.10
& 3.20 $\pm$ 0.10
& 15.24 $\pm$ 0.00
\\

\bottomrule
\end{tabular}

\vspace{0.5em}

\begin{minipage}{0.95\linewidth}
\footnotesize
\textbf{Metrics:}
Runtime corresponds to mean wall-clock inference time per registration pair.
Throughput denotes registrations processed per second and is derived as the inverse runtime.
CPU usage is normalized over all logical CPU cores.
GPU peak corresponds to maximum allocated GPU memory during inference.
\end{minipage}

\end{table}

\section{Discussion}
\label{sec:Discussion}
Multimodal deformable image registration remains a highly challenging problem. Unlike monomodal registration, multimodal registration cannot rely on direct intensity correspondence and must instead infer spatial relationships between images that encode the same anatomy through fundamentally different physical contrast mechanisms. Consequently, the registration quality depends simultaneously on the transformation model, similarity measure, regularization strategy, anatomical region and modality pair. The results of this benchmark demonstrate that no evaluated method achieves consistently superior performance across all datasets and evaluation criteria, and that registration in a real-world scenario with limited training data remains a challenging and unsolved problem in many clinically relevant settings..

A first major observation is the strong dependence of registration performance on anatomical region and modality pair. Classical methods exhibit particularly heterogeneous behavior. ANTs SyN achieves moderate improvements on BRATS but degrades performance on HCP, whereas Elastix performs competitively on the synthetic brain benchmarks but deteriorates substantially on IMAG2 (Pelvis). This variability suggests that handcrafted multimodal similarity measures remain highly sensitive to the appearance relationship between modalities and to the anatomical context. In this specific case, the low visibility of the skull in the b0 of the HCP dataset is assumed to be the cause of such a failure for ANTs SyN. By contrast, learning-based approaches provide more consistent performance across datasets, with GradICON and VoxelMorph generally outperforming classical methods on BRATS and HCP. However, even the best-performing methods exhibit only limited to no improvement on the IMAG2 (pelvis) dataset, indicating that robust multimodal registration remains highly dependent on the clinical setup.

A second and perhaps more surprising finding concerns the relationship between deformation recovery and registration accuracy. On the synthetic BRATS and HCP benchmarks, the availability of ground-truth deformation fields made it possible to evaluate both the quality of the resulting registration and the accuracy with which the original deformation was recovered. First of all, for both classical and learning-based methods, the DVF percentage error remains over 80\% and goes up to 211\%, which outlines a poor recovery capability. DVF errors remain high, indicating that the recovered deformations differ substantially from the synthetic fields used for data generation. However, because image registration is inherently non-identifiable, deformation recovery should not be interpreted as a direct measure of registration accuracy. In addition, the results reveal that these two objectives are not equivalent. In several cases, methods achieving the lowest DVF errors did not produce the highest $\overline{\mathrm{DSC}}$ or lowest $\overline{\mathrm{HD95}}$ values. For example, M2M-GradICON consistently achieved among the lowest deformation recovery errors while remaining substantially inferior to GradICON in terms of overlap-based metrics. Conversely, GradICON and VoxelMorph frequently obtained superior $\overline{\mathrm{DSC}}$, $\overline{\mathrm{HD95}}$, and MIND-SSC loss values while exhibiting larger deformation errors. This observation suggests that a deformation field capable of producing optimal anatomical correspondence according to registration metrics may differ significantly from the synthetic deformation used to generate the benchmark. More broadly, these results indicate that synthetic deformation recovery and multimodal anatomical registration should be regarded as related but distinct tasks.

A third major observation is the trade-off between registration accuracy and transformation regularity. The folding-ratio results demonstrate that the methods producing the largest gains in overlap and boundary alignment are often those generating the largest number of local topology violations (although the maximum occurrence remains very limited). In particular, VoxelMorph and GradICON achieved some of the strongest registration performances on BRATS and HCP but also produced the highest folding ratios. Conversely, M2M-GradICON, multiGradICON, and the classical methods generated nearly fold-free transformations, albeit at the expense of reduced registration accuracy. This trade-off highlights a fundamental challenge in deformable registration. Maximizing image alignment alone can encourage transformations that are anatomically implausible, while strong regularization can limit the ability of the model to match complex deformations. Since local folding corresponds to non-invertible mappings that cannot be interpreted as physically meaningful anatomical transformations, deformation regularity should be considered a primary evaluation criterion rather than a secondary quality measure.

The anatomy-guided experiments provide additional insight into the behavior of multimodal registration algorithms. Incorporating segmentation information dramatically improves overlap-based metrics, with ANTs SyN seg.\ achieving $\overline{\mathrm{DSC}}$ values above 90\% on BRATS and above 80\% on both HCP and IMAG2 (Pelvis). However, these improvements do not translate consistently to the remaining evaluation criteria. In particular, anatomy-guided approaches often exhibit substantially worse MIND-SSC loss values than their unsupervised counterparts and do not systematically improve deformation recovery. This observation suggests that anatomy-guided registration should not be interpreted simply as a more accurate version of unsupervised registration. Instead, anatomical supervision changes the optimization objective itself by explicitly prioritizing agreement on annotated structures. Such a behavior may be desirable in applications focusing on contour propagation, treatment planning, or organ tracking. However, it also indicates that high segmentation overlap does not necessarily imply improved global multimodal correspondence throughout the image volume.

The comparison between ANTs SyN seg.\ and GradICON seg.\ further highlights this distinction. ANTs SyN seg.\ consistently achieves the highest $\overline{\mathrm{DSC}}$ values and among the lowest $\overline{\mathrm{HD95}}$ values, demonstrating its ability to maximize alignment of labeled structures. By contrast, GradICON seg.\ systematically produces substantially lower MIND-SSC loss and, on BRATS, markedly better deformation recovery. These results suggest that the two methods optimize different notions of registration quality. Whereas ANTs SyN seg.\ focuses on matching annotated structures as accurately as possible, GradICON seg.\ appears to preserve a broader notion of multimodal structural consistency. This is a direct consequence of the training protocol used for GradICON seg. \ which combined both segmentation overlap and MIND-SSC loss value in its optimization criterion whereas ANTs SyN seg. \ only focused on segmentation overlap. This distinction reinforces the importance of evaluating anatomy-guided methods using multiple complementary metrics rather than overlap measures alone.

The behavior of multiGradICON deserves particular attention. As a foundation model trained on a large and diverse collection of registration tasks, one might expect it to generalize more effectively than task-specific approaches. However, the results demonstrate that multiGradICON does not provide consistent performance across datasets and neither consistently outperform the smaller GradICON models. In fact, it is frequently inferior in terms of $\overline{\mathrm{DSC}}$, $\overline{\mathrm{HD95}}$, and $\mathcal{L}_{MIND-SSC}$, particularly on IMAG2 (pelvis). Several explanations are possible. First, the model was evaluated in a zero-shot setting without dataset-specific adaptation. 
Secondly, the training objective of multiGradICON emphasizes transformation consistency and robustness across tasks rather than optimization for any single registration benchmark. Regardless of the underlying cause, these results indicate that large-scale pretraining alone does not currently guarantee robust multimodal registration performance across anatomies and modality pairs.

Perhaps the most clinically relevant finding of this benchmark is the discrepancy between performance on synthetic and real registration tasks. On BRATS and HCP, multiple methods substantially improve registration accuracy across most metrics. By contrast, on IMAG2 (Pelvis), virtually no unsupervised method achieves a meaningful improvement over the unregistered baseline. The best-performing methods do not modify the $\overline{\mathrm{DSC}}$ and $\overline{\mathrm{HD95}}$ at all despite the use of state-of-the-art registration frameworks. This contrast suggests that current synthetic benchmarks may overestimate the capability of registration algorithms to solve real multimodal correspondence problems. While synthetic deformations provide a valuable controlled evaluation setting, they cannot reproduce all the uncertainty associated with anatomical variability, modality-specific visibility, partial volume effects, acquisition artifacts, and ambiguous correspondences encountered in clinical practice. These findings therefore highlight the need for larger collections of real multimodal datasets with expert annotations and clinically meaningful evaluation criteria.

The benchmark also exposes a broader limitation of multimodal registration research: the absence of dense ground-truth correspondences. Each available evaluation strategy captures only a part of the problem. The segmentation overlap evaluates selected structures but ignores unlabeled anatomy. DVF recovery is only available for synthetic experiments and may not correlate with registration performance. Image-based similarity metrics such as MIND-SSC loss provide dense measurements but do not necessarily reflect clinically relevant structure alignment. Topology-preservation metrics quantify transformation plausibility but not registration accuracy. Consequently, the registration quality cannot be characterized by any single metric. The disagreement observed throughout this benchmark strongly supports the adoption of multi-criteria evaluation protocols incorporating geometric, photometric, deformation-based, and anatomical measures simultaneously.

An additional challenge revealed by this benchmark concerns the interpretation of quantitative metrics themselves. Although numerical evaluation is essential for objective comparison, quantitative scores do not always reflect the perceived quality or practical usefulness of a registration. Across several datasets, visual inspection often suggested a more favorable alignment than indicated by the reported metrics. For example, in BRATS, the complex and irregular morphology of brain tumors can lead to moderate $\overline{\mathrm{DSC}}$ values despite visually satisfactory alignment of the tumor boundaries. Similarly, the highly folded cortical anatomy can amplify local discrepancies captured by image-based similarity metrics such as MIND-SSC, even when the overall anatomical correspondence appears clinically acceptable. In HCP, the problem is further complicated by modality-specific visibility differences. Since structures outside the brain, including portions of the skull and facial anatomy, are hardly visible on $b_0$ images, registration methods may introduce substantial deformations in these regions that negatively affect quantitative measurements while having limited impact on the clinically relevant intracranial alignment. Finally, in IMAG2 (Pelvis), the lower image quality of the fixed image and the difficulty of establishing unambiguous multimodal correspondences result in only small visual differences between methods despite measurable quantitative variations. Depending on the intended application, such discrepancies may or may not be clinically meaningful.

Several limitations of the present study should nevertheless be acknowledged. First, the brain experiments rely on synthetically generated deformations and therefore evaluate the recovery of known transformations rather than true multimodal correspondence estimation. Secondly, the benchmark remains dependent on the availability and quality of anatomical segmentations. Thirdly, although for classical methods we performed a dataset-specific hyperparameter tuning, learning-based approaches were evaluated using the hyperparameters proposed by their authors. This choice was made because performing a comparable grid search would have been prohibitively time-consuming. As a result, some performance differences may reflect hyperparameter sensitivity rather than intrinsic methodological superiority. Finally, the IMAG2 (pelvis) dataset remains relatively small and does not provide dense correspondence annotations, preventing a precise evaluation of voxel-level registration accuracy.

Future progress in multimodal deformable registration will likely require advances beyond similarity metrics alone. In particular, the acquisition of large real-world multimodal datasets will probably determine the outcome of future research. Specific focus should be laid on providing datasets containing genuinely aligned multimodal acquisitions (through animal or human acquisitions on sedated patients) and segmentation labels distributed throughout the whole field of view to maximize the correspondence between the anatomical and similarity metrics. On a similar level, evaluation protocols tied to downstream clinical tasks may be an alternate solution to assess clinical validity of the registration strategies, as it helps to tie back the registration performance to a single relevant metric. On a second level, improved anatomical and biomechanical priors and stronger topology-preserving optimization strategies represent promising directions. More generally, the results of this benchmark suggest that future methods should not optimize exclusively for overlap accuracy or image similarity, but instead seek a balanced compromise between anatomical correspondence, multimodal consistency, accurate deformation estimation, and transformation plausibility.

\section{Conclusion}
\label{sec:Conclusion}
Overall, the results of this benchmark demonstrate that multimodal deformable registration remains an open and highly context-dependent problem. Anatomy-guided methods can achieve excellent alignment of labeled structures, but may not guarantee global multimodal consistency. Learning-based methods such as GradICON and VoxelMorph provide competitive and sometimes more balanced performance, but their robustness across anatomical regions and real clinical deformation scenarios remains limited. Classical methods remain useful baselines but show strong dataset dependence. These findings support the need for comprehensive, multi-metric evaluation and suggest that clinically reliable multimodal registration will require methods that jointly address anatomical accuracy, image-based consistency, physically plausible deformation and downstream clinical task performance.

\section*{Acknowledgments}
This work has been funded and supported by Ligue contre le cancer. 

The data for the IMAG2 (Pelvis) dataset was acquired following the ClinicalTrials ID NCT06224985, APHP ID 2015-A01705-44.

This project was provided with computing AI and storage resources by GENCI at IDRIS thanks to the grant 2025-AD010316730 on the supercomputer Jean Zay V100 and A100 partitions.

\bibliographystyle{elsarticle-num} 
\bibliography{references}

@article{2002Maes,
  title={Multimodality image registration by maximization of mutual information},
  author={Maes, Frederik and Collignon, Andre and Vandermeulen, Dirk and Marchal, Guy and Suetens, Paul},
  journal={IEEE transactions on Medical Imaging},
  volume={16},
  number={2},
  pages={187--198},
  year={2002},
  publisher={IEEE}
}

@article{2002Rueckert,
  title={Nonrigid registration using free-form deformations: application to breast {MR} images},
  author={Rueckert, Daniel and Sonoda, Luke I and Hayes, Carmel and Hill, Derek LG and Leach, Martin O and Hawkes, David J},
  journal={IEEE {T}ransactions on {M}edical {I}maging},
  volume={18},
  number={8},
  pages={712--721},
  year={2002},
  publisher={IEEE}
}

@article{2005Beg,
  title={Computing large deformation metric mappings via geodesic flows of diffeomorphisms},
  author={Beg, M. Faisal and Miller, Michael I and Trouv{\'e}, Alain and Younes, Laurent},
  journal={International {J}ournal of {C}omputer {V}ision},
  volume={61},
  number={2},
  pages={139--157},
  year={2005},
  publisher={Springer}
}

@article{2006Crum,
  title={Generalized overlap measures for evaluation and validation in medical image analysis},
  author={Crum, William R and Camara, Oscar and Hill, Derek LG},
  journal={IEEE {T}ransactions on {M}edical {I}maging},
  volume={25},
  number={11},
  pages={1451--1461},
  year={2006},
  publisher={IEEE}
}

@inproceedings{2006Haber,
  title={Intensity gradient based registration and fusion of multi-modal images},
  author={Haber, Eldad and Modersitzki, Jan},
  booktitle={International Conference on Medical Image Computing and Computer-Assisted Intervention},
  pages={726--733},
  year={2006},
  organization={Springer}
}

@article{2008Avants,
  title={Symmetric diffeomorphic image registration with cross-correlation: evaluating automated labeling of elderly and neurodegenerative brain},
  author={Avants, Brian B and Epstein, Charles L and Grossman, Murray and Gee, James C},
  journal={Medical {I}mage {A}nalysis},
  volume={12},
  number={1},
  pages={26--41},
  year={2008},
  publisher={Elsevier}
}

@article{2009Klein,
  title={Elastix: a toolbox for intensity-based medical image registration},
  author={Klein, Stefan and Staring, Marius and Murphy, Keelin and Viergever, Max A and Pluim, Josien PW},
  journal={IEEE {T}ransactions on {M}edical {I}maging},
  volume={29},
  number={1},
  pages={196--205},
  year={2009},
  publisher={IEEE}
}

@article{2009Vercauteren,
  title={Diffeomorphic demons: Efficient non-parametric image registration},
  author={Vercauteren, Tom and Pennec, Xavier and Perchant, Aymeric and Ayache, Nicholas},
  journal={NeuroImage},
  volume={45},
  number={1},
  pages={S61--S72},
  year={2009},
  publisher={Elsevier}
}

@article{2012Heinrich,
  title={{MIND}: {M}odality {I}ndependent {N}eighbourhood {D}escriptor for multi-modal deformable registration},
  author={Heinrich, Mattias P and Jenkinson, Mark and Bhushan, Manav and Matin, Tahreema and Gleeson, Fergus V and Brady, Michael and Schnabel, Julia A},
  journal={Medical {I}mage {A}nalysis},
  volume={16},
  number={7},
  pages={1423--1435},
  year={2012},
  publisher={Elsevier}
}

@article{2013Glasser,
  title={The minimal preprocessing pipelines for the {H}uman {C}onnectome {P}roject},
  author={Glasser, Matthew F and Sotiropoulos, Stamatios N and Wilson, J Anthony and Coalson, Timothy S and Fischl, Bruce and Andersson, Jesper L and Xu, Junqian and Jbabdi, Saad and Webster, Matthew and Polimeni, Jonathan R and others},
  journal={Neuroimage},
  volume={80},
  pages={105--124},
  year={2013},
  publisher={Elsevier}
}

@inproceedings{2013Heinrich,
  title={Towards realtime multimodal fusion for image-guided interventions using self-similarities},
  author={Heinrich, Mattias Paul and Jenkinson, Mark and Papie{\.z}, Bartlomiej W and Brady, Sir Michael and Schnabel, Julia A},
  booktitle={International {C}onference on {M}edical {I}mage {C}omputing and {C}omputer-{A}ssisted {I}ntervention},
  pages={187--194},
  year={2013},
  organization={Springer}
}

@article{2013Sotiras,
  title={Deformable medical image registration: {A} survey},
  author={Sotiras, Aristeidis and Davatzikos, Christos and Paragios, Nikos},
  journal={IEEE {T}ransactions on {M}edical {I}maging},
  volume={32},
  number={7},
  pages={1153--1190},
  year={2013},
  publisher={IEEE}
}

@article{2013VanEssen,
  title={The {W}u-{M}inn {H}uman {C}onnectome {P}roject: an overview},
  author={Van Essen, David C and Smith, Stephen M and Barch, Deanna M and Behrens, Timothy EJ and Yacoub, Essa and Ugurbil, Kamil and Wu-Minn HCP Consortium and others},
  journal={Neuroimage},
  volume={80},
  pages={62--79},
  year={2013},
  publisher={Elsevier}
}

@article{2014Lombaert,
  title={Spectral log-demons: diffeomorphic image registration with very large deformations},
  author={Lombaert, Herve and Grady, Leo and Pennec, Xavier and Ayache, Nicholas and Cheriet, Farida},
  journal={International {J}ournal of {C}omputer {V}ision},
  volume={107},
  number={3},
  pages={254--271},
  year={2014},
  publisher={Springer}
}

@article{2014Menze,
  title={The multimodal {BRA}in {T}umor image {S}egmentation benchmark ({BRATS})},
  author={Menze, Bjoern H and Jakab, Andras and Bauer, Stefan and Kalpathy-Cramer, Jayashree and Farahani, Keyvan and Kirby, Justin and Burren, Yuliya and Porz, Nicole and Slotboom, Johannes and Wiest, Roland and others},
  journal={IEEE {T}ransactions on {M}edical {I}maging},
  volume={34},
  number={10},
  pages={1993--2024},
  year={2014},
  publisher={IEEE}
}

@article{2015Sudlow,
  title={{UK} {B}iobank: an open access resource for identifying the causes of a wide range of complex diseases of middle and old age},
  author={Sudlow, Cathie and Gallacher, John and Allen, Naomi and Beral, Valerie and Burton, Paul and Danesh, John and Downey, Paul and Elliott, Paul and Green, Jane and Landray, Martin and others},
  journal={PLoS medicine},
  volume={12},
  number={3},
  pages={e1001779},
  year={2015},
  publisher={Public Library of Science}
}

@article{2015Tilak,
  title={3{T} {MR}-guided in-bore transperineal prostate biopsy: {A} comparison of robotic and manual needle-guidance templates},
  author={Tilak, Gaurie and Tuncali, Kemal and Song, Sang-Eun and Tokuda, Junichi and Olubiyi, Olutayo and Fennessy, Fiona and Fedorov, Andriy and Penzkofer, Tobias and Tempany, Clare and Hata, Nobuhiko},
  journal={Journal of Magnetic Resonance Imaging},
  volume={42},
  number={1},
  pages={63--71},
  year={2015},
  publisher={Wiley Online Library}
}

@inproceedings{2016simonovsky,
  title={A deep metric for multimodal registration},
  author={Simonovsky, Martin and Guti{\'e}rrez-Becker, Benjam{\'\i}n and Mateus, Diana and Navab, Nassir and Komodakis, Nikos},
  booktitle={International conference on medical image computing and computer-assisted intervention},
  pages={10--18},
  year={2016},
  organization={Springer}
}

@article{2017Brock,
  title={Use of image registration and fusion algorithms and techniques in radiotherapy: Report of the AAPM Radiation Therapy Committee Task Group No. 132},
  author={Brock, Kristy K and Mutic, Sasa and McNutt, Todd R and Li, Hua and Kessler, Marc L},
  journal={Medical physics},
  volume={44},
  number={7},
  pages={e43--e76},
  year={2017},
  publisher={Wiley Online Library}
}

@article{2017Jiang,
  title={Multimodal image registration based on binary gradient angle descriptor},
  author={Jiang, Dongsheng and Shi, Yonghong and Yao, Demin and Fan, Yifeng and Wang, Manning and Song, Zhijian},
  journal={International Journal of Computer Assisted Radiology and Surgery},
  volume={12},
  number={12},
  pages={2157--2167},
  year={2017},
  publisher={Springer}
}

@article{2018Hu,
  title={Weakly-supervised convolutional neural networks for multimodal image registration},
  author={Hu, Yipeng and Modat, Marc and Gibson, Eli and Li, Wenqi and Ghavami, Nooshin and Bonmati, Ester and Wang, Guotai and Bandula, Steven and Moore, Caroline M and Emberton, Mark and others},
  journal={Medical {I}mage {A}nalysis},
  volume={49},
  pages={1--13},
  year={2018},
  publisher={Elsevier}
}

@article{2019Balakrishnan,
  title={Voxelmorph: a learning framework for deformable medical image registration},
  author={Balakrishnan, Guha and Zhao, Amy and Sabuncu, Mert R and Guttag, John and Dalca, Adrian V},
  journal={IEEE {T}ransactions on {M}edical {I}maging},
  volume={38},
  number={8},
  pages={1788--1800},
  year={2019},
  publisher={IEEE}
}

@article{2019Fan,
  title={Adversarial learning for mono-or multi-modal registration},
  author={Fan, Jingfan and Cao, Xiaohuan and Wang, Qian and Yap, Pew-Thian and Shen, Dinggang},
  journal={Medical {I}mage {A}nalysis},
  volume={58},
  pages={101545},
  year={2019},
  publisher={Elsevier}
}

@article{2019Haskins,
  title={Learning deep similarity metric for {3D} {MR}--{TRUS} image registration},
  author={Haskins, Grant and Kruecker, Jochen and Kruger, Uwe and Xu, Sheng and Pinto, Peter A and Wood, Brad J and Yan, Pingkun},
  journal={International {J}ournal of {C}omputer {A}ssisted {R}adiology and {S}urgery},
  volume={14},
  number={3},
  pages={417--425},
  year={2019},
  publisher={Springer}
}

@inproceedings{2019Niethammer,
  title={Metric learning for image registration},
  author={Niethammer, Marc and Kwitt, Roland and Vialard, Francois-Xavier},
  booktitle={2019 IEEE/CVF Conference on Computer Vision and Pattern Recognition (CVPR)},
  pages={8455--8464},
  year={2019},
  organization={IEEE}
}

@article{2019Tournier,
  title={{MR}trix3: A fast, flexible and open software framework for medical image processing and visualisation},
  author={Tournier, J-Donald and Smith, Robert and Raffelt, David and Tabbara, Rami and Dhollander, Thijs and Pietsch, Maximilian and Christiaens, Daan and Jeurissen, Ben and Yeh, Chun-Hung and Connelly, Alan},
  journal={Neuroimage},
  volume={202},
  pages={116137},
  year={2019},
  publisher={Elsevier}
}

@inproceedings{2019Wei,
  title={Synthesis and inpainting-based {MR}-{CT} registration for image-guided thermal ablation of liver tumors},
  author={Wei, Dongming and Ahmad, Sahar and Huo, Jiayu and Peng, Wen and Ge, Yunhao and Xue, Zhong and Yap, Pew-Thian and Li, Wentao and Shen, Dinggang and Wang, Qian},
  booktitle={International {C}onference on {M}edical {I}mage {C}omputing and {C}omputer-{A}ssisted {I}ntervention},
  pages={512--520},
  year={2019},
  organization={Springer}
}

@article{2019Dalca,
  title={Unsupervised learning of probabilistic diffeomorphic registration for images and surfaces},
  author={Dalca, Adrian V and Balakrishnan, Guha and Guttag, John and Sabuncu, Mert R},
  journal={Medical {I}mage {A}nalysis},
  volume={57},
  pages={226--236},
  year={2019},
  publisher={Elsevier}
}

@article{2019DeVos,
  title={A deep learning framework for unsupervised affine and deformable image registration},
  author={De Vos, Bob D and Berendsen, Floris F and Viergever, Max A and Sokooti, Hessam and Staring, Marius and I{\v{s}}gum, Ivana},
  journal={Medical {I}mage {A}nalysis},
  volume={52},
  pages={128--143},
  year={2019},
  publisher={Elsevier}
}

@article{2020Fu,
  title={Deep learning in medical image registration: {A} review},
  author={Fu, Yabo and Lei, Yang and Wang, Tonghe and Curran, Walter J and Liu, Tian and Yang, Xiaofeng},
  journal={Physics in Medicine \& Biology},
  volume={65},
  number={20},
  pages={20TR01},
  year={2020},
  publisher={IOP Publishing}
}

@article{2020Haskins,
  title={Deep learning in medical image registration: a survey},
  author={Haskins, Grant and Kruger, Uwe and Yan, Pingkun},
  journal={Machine Vision and Applications},
  volume={31},
  number={1},
  pages={8},
  year={2020},
  publisher={Springer}
}

@article{2020Pielawski,
  title={Co{MIR}: {C}ontrastive {M}ultimodal {I}mage {R}epresentation for registration},
  author={Pielawski, Nicolas and Wetzer, Elisabeth and {\"O}fverstedt, Johan and Lu, Jiahao and W{\"a}hlby, Carolina and Lindblad, Joakim and Sladoje, Natasa},
  journal={Advances in neural information processing systems},
  volume={33},
  pages={18433--18444},
  year={2020}
}

@inproceedings{2020Xu,
  title={Adversarial uni-and multi-modal stream networks for multimodal image registration},
  author={Xu, Zhe and Luo, Jie and Yan, Jiangpeng and Pulya, Ritvik and Li, Xiu and Wells III, William and Jagadeesan, Jayender},
  booktitle={International Conference on Medical Image Computing and Computer-Assisted Intervention},
  pages={222--232},
  year={2020},
  organization={Springer}
}

@article{2021bChen,
  title={Vit-{V}-{N}et: {V}ision transformer for unsupervised volumetric medical image registration},
  author={Chen, Junyu and He, Yufan and Frey, Eric C and Li, Ye and Du, Yong},
  journal={arXiv preprint arXiv:2104.06468},
  year={2021}
}

@inproceedings{2021czolbe,
  title={Semantic similarity metrics for learned image registration},
  author={Czolbe, Steffen and Krause, Oswin and Feragen, Aasa},
  booktitle={Medical Imaging with Deep Learning},
  pages={105--118},
  year={2021},
  organization={PMLR}
}

@article{2021Kim,
  title={Cycle{M}orph: cycle consistent unsupervised deformable image registration},
  author={Kim, Boah and Kim, Dong Hwan and Park, Seong Ho and Kim, Jieun and Lee, June-Goo and Ye, Jong Chul},
  journal={Medical {I}mage {A}nalysis},
  volume={71},
  pages={102036},
  year={2021},
  publisher={Elsevier}
}

@article{2021Kong,
  title={Breaking the dilemma of medical image-to-image translation},
  author={Kong, Lingke and Lian, Chenyu and Huang, Detian and Hu, Yanle and Zhou, Qichao and others},
  journal={Advances in Neural Information Processing Systems},
  volume={34},
  pages={1964--1978},
  year={2021}
}

@inproceedings{2021Liu,
  title={{SAME}: Deformable image registration based on self-supervised anatomical embeddings},
  author={Liu, Fengze and Yan, Ke and Harrison, Adam P and Guo, Dazhou and Lu, Le and Yuille, Alan L and Huang, Lingyun and Xie, Guotong and Xiao, Jing and Ye, Xianghua and others},
  booktitle={International Conference on Medical Image Computing and Computer-Assisted Intervention},
  pages={87--97},
  year={2021},
  organization={Springer}
}

@article{2021Lu,
  title={Is image-to-image translation the panacea for multimodal image registration? {A} comparative study},
  author={Lu, Jiahao and others},
  journal={Medical Image Analysis},
  year={2021}
}

@article{2021Hu,
  title={End-to-end multimodal image registration via reinforcement learning},
  author={Hu, Jing and Luo, Ziwei and Wang, Xin and Sun, Shanhui and Yin, Youbing and Cao, Kunlin and Song, Qi and Lyu, Siwei and Wu, Xi},
  journal={Medical {I}mage {A}nalysis},
  volume={68},
  pages={101878},
  year={2021},
  publisher={Elsevier}
}

@article{2021Xiao,
  title={A review of deep learning-based three-dimensional medical image registration methods},
  author={Xiao, Haonan and Teng, Xinzhi and Liu, Chenyang and Li, Tian and Ren, Ge and Yang, Ruijie and Shen, Dinggang and Cai, Jing},
  journal={Quantitative Imaging in Medicine and Surgery},
  volume={11},
  number={12},
  pages={4895},
  year={2021}
}

@article{2022Chen,
  title={Trans{M}orph: {T}ransformer for unsupervised medical image registration},
  author={Chen, Junyu and Frey, Eric C and He, Yufan and Segars, William P and Li, Ye and Du, Yong},
  journal={Medical {I}mage {A}nalysis},
  volume={82},
  pages={102615},
  year={2022},
  publisher={Elsevier}
}

@inproceedings{2022Dey,
  title={Contra{R}eg: Contrastive learning of multi-modality unsupervised deformable image registration},
  author={Dey, Neel and Schlemper, Jo and Salehi, Seyed Sadegh Mohseni and Zhou, Bo and Gerig, Guido and Sofka, Michal},
  booktitle={International Conference on Medical Image Computing and Computer-Assisted Intervention},
  pages={66--77},
  year={2022},
  organization={Springer}
}

@article{2022Hering,
  title={Learn{2R}eg: comprehensive multi-task medical image registration challenge, dataset and evaluation in the era of deep learning},
  author={Hering, Alessa and Hansen, Lasse and Mok, Tony CW and Chung, Albert CS and Siebert, Hanna and H{\"a}ger, Stephanie and Lange, Annkristin and Kuckertz, Sven and Heldmann, Stefan and Shao, Wei and others},
  journal={IEEE Transactions on Medical Imaging},
  volume={42},
  number={3},
  pages={697--712},
  year={2022},
  publisher={IEEE}
}

@inproceedings{2022Jian,
  title={Weakly-supervised biomechanically-constrained {CT}/{MRI} registration of the spine},
  author={Jian, Bailiang and Azampour, Mohammad Farid and De Benetti, Francesca and Oberreuter, Johannes and Bukas, Christina and Gersing, Alexandra S and Foreman, Sarah C and Dietrich, Anna-Sophia and Rischewski, Jon and Kirschke, Jan S and others},
  booktitle={International Conference on Medical Image Computing and Computer-Assisted Intervention},
  pages={227--236},
  year={2022},
  organization={Springer}
}

@article{2022Song,
  title={Cross-modal attention for multi-modal image registration},
  author={Song, Xinrui and Chao, Hanqing and Xu, Xuanang and Guo, Hengtao and Xu, Sheng and Turkbey, Baris and Wood, Bradford J and Sanford, Thomas and Wang, Ge and Yan, Pingkun},
  journal={Medical Image Analysis},
  volume={82},
  pages={102612},
  year={2022},
  publisher={Elsevier}
}

@article{2022Zou,
  title={A review of deep learning-based deformable medical image registration},
  author={Zou, Jing and Gao, Bingchen and Song, Youyi and Qin, Jing},
  journal={Frontiers in Oncology},
  volume={12},
  pages={1047215},
  year={2022},
  publisher={Frontiers Media SA}
}

@article{2023Adewole,
  title={The {BRA}in {T}umor {S}egmentation ({BRATS}) challenge 2023: Glioma segmentation in sub-saharan africa patient population ({BRATS}-africa)},
  author={Adewole, Maruf and Rudie, Jeffrey D and Gbdamosi, Anu and Toyobo, Oluyemisi and Raymond, Confidence and Zhang, Dong and Omidiji, Olubukola and Akinola, Rachel and Suwaid, Mohammad Abba and Emegoakor, Adaobi and others},
  journal={ArXiv},
  pages={arXiv--2305},
  year={2023}
}

@article{2023Liu,
  title={Geometry-consistent adversarial registration model for unsupervised multi-modal medical image registration},
  author={Liu, Yanxia and Wang, Wenqi and Li, Yuhong and Lai, Haoyu and Huang, Sijuan and Yang, Xin},
  journal={IEEE Journal of Biomedical and Health Informatics},
  volume={27},
  number={7},
  pages={3455--3466},
  year={2023},
  publisher={IEEE}
}

@inproceedings{2023Ronchetti,
  title={{DISA}: {DI}fferentiable {S}imilarity {A}pproximation for universal multimodal registration},
  author={Ronchetti, Matteo and Wein, Wolfgang and Navab, Nassir and Zettinig, Oliver and Prevost, Raphael},
  booktitle={International Conference on Medical Image Computing and Computer-Assisted Intervention},
  pages={761--770},
  year={2023},
  organization={Springer}
}

@inproceedings{2023Tian,
  title={Grad{ICON}: Approximate diffeomorphisms via gradient inverse consistency},
  author={Tian, Lin and Greer, Hastings and Vialard, Fran{\c{c}}ois-Xavier and Kwitt, Roland and Est{\'e}par, Ra{\'u}l San Jos{\'e} and Rushmore, Richard Jarrett and Makris, Nikolaos and Bouix, Sylvain and Niethammer, Marc},
  booktitle={Proceedings of the IEEE/CVF Conference on Computer Vision and Pattern Recognition},
  pages={18084--18094},
  year={2023}
}

@article{2023Wang,
  title={Multimodal registration of ultrasound and mr images using weighted self-similarity structure vector},
  author={Wang, Yifan and Fu, Tianyu and Wu, Chan and Xiao, Jian and Fan, Jingfan and Song, Hong and Liang, Ping and Yang, Jian},
  journal={Computers in Biology and Medicine},
  volume={155},
  pages={106661},
  year={2023},
  publisher={Elsevier}
}

@inproceedings{2024Demir,
  title={multi{G}rad{ICON}: {A} foundation model for multimodal medical image registration},
  author={Demir, Ba{\c{s}}ar and Tian, Lin and Greer, Hastings and Kwitt, Roland and Vialard, Fran{\c{c}}ois-Xavier and Est{\'e}par, Ra{\'u}l San Jos{\'e} and Bouix, Sylvain and Rushmore, Richard and Ebrahim, Ebrahim and Niethammer, Marc},
  booktitle={International Workshop on Biomedical Image Registration},
  pages={3--18},
  year={2024},
  organization={Springer}
}

@article{2024Gao,
  title={{MAIRN}et: weakly supervised anatomy-aware multimodal articulated image registration network},
  author={Gao, Xiaoru and Zhong, Woquan and Wang, Runze and Heimann, Alexander F and Tannast, Moritz and Zheng, Guoyan},
  journal={International Journal of Computer Assisted Radiology and Surgery},
  volume={19},
  number={3},
  pages={507--517},
  year={2024},
  publisher={Springer}
}

@inproceedings{2024Mok,
  title={Modality-agnostic structural image representation learning for deformable multi-modality medical image registration},
  author={Mok, Tony CW and Li, Zi and Bai, Yunhao and Zhang, Jianpeng and Liu, Wei and Zhou, Yan-Jie and Yan, Ke and Jin, Dakai and Shi, Yu and Yin, Xiaoli and others},
  booktitle={Proceedings of the IEEE/CVF Conference on Computer Vision and Pattern Recognition},
  pages={11215--11225},
  year={2024}
}

@article{2024Ramadan,
  title={Medical image registration in the era of Transformers: {A} recent review},
  author={Ramadan, Hiba and El Bourakadi, Dounia and Yahyaouy, Ali and Tairi, Hamid},
  journal={Informatics in Medicine Unlocked},
  volume={49},
  pages={101540},
  year={2024},
  publisher={Elsevier}
}

@article{2025Chenb,
  title={A survey on deep learning in medical image registration: New technologies, uncertainty, evaluation metrics, and beyond},
  author={Chen, Junyu and Liu, Yihao and Wei, Shuwen and Bian, Zhangxing and Subramanian, Shalini and Carass, Aaron and Prince, Jerry L and Du, Yong},
  journal={Medical {I}mage {A}nalysis},
  volume={100},
  pages={103385},
  year={2025},
  publisher={Elsevier}
}

@inproceedings{2025Choo,
  title={Mono-{M}odalizing extremely heterogeneous multi-modal medical image registration},
  author={Choo, Kyobin and Han, Hyunkyung and Kim, Jinyeong and Yoon, Chanyong and Hwang, Seong Jae},
  booktitle={International Conference on Medical Image Computing and Computer-Assisted Intervention},
  pages={433--443},
  year={2025},
  organization={Springer}
}

@article{2025Hansen,
  title={Learn{2R}eg 2024: New Benchmark Datasets Driving Progress on New Challenges},
  author={Hansen, Lasse and Heyer, Wiebke and Gro{\ss}br{\"o}hmer, Christoph and Madesta, Frederic and Sentker, Thilo and Jiazheng, Wang and Zhang, Yuxi and Zhang, Hang and Liu, Min and Wang, Junyi and others},
  journal={arXiv preprint arXiv:2509.01217},
  year={2025}
}

@article{2025He,
  title={Match{A}nything: Universal Cross-Modality Image Matching with Large-Scale Pre-Training. arXiv 2025},
  author={He, X and Yu, H and Peng, S and Tan, D and Shen, Z and Bao, H and Zhou, X},
  journal={arXiv preprint arXiv:2501.07556},
  year={2025}
}

@article{2025bHe,
  title={{SAMIR}, an efficient registration framework via robust feature learning from {SAM}},
  author={He, Yue and Liu, Min and Liu, Qinghao and Wang, Jiazheng and Wang, Yaonan and Zhang, Hang and Chen, Xiang},
  journal={arXiv preprint arXiv:2509.13629},
  year={2025}
}

@inproceedings{2025Tursynbek,
  title={Guiding registration with emergent similarity from pre-trained diffusion models},
  author={Tursynbek, Nurislam and Greer, Hastings and Demir, Ba{\c{s}}ar and Niethammer, Marc},
  booktitle={International Conference on Medical Image Computing and Computer-Assisted Intervention},
  pages={240--251},
  year={2025},
  organization={Springer}
}

@article{2025Wang,
  title={Weakly Supervised Spatial Implicit Neural Representation Learning for {3D} {MRI}-{U}ltrasound Deformable Image Registration in {HDR} Prostate Brachytherapy: {TU}-130-202-10},
  author={Wang, Jing and Liu, Ruirui and Lei, Yu and Baine, Michael J. and Liu, Tian},
  journal={Medical Physics},
  volume={52},
  number={10},
  pages={e70059527--e70059528},
  year={2025}
}

@inproceedings{2026Nascimento,
  title={Image Synthesis and Transfer Learning for Estimating Multi-Modal Brain Registration Error},
  author={Nascimento, Leandro and Fran{\c{c}}ois, Quentin and Duplat, Bertrand and Haliyo, Sinan and Bloch, Isabelle},
  booktitle={2026 IEEE 23rd International Symposium on Biomedical Imaging (ISBI)},
  pages={1--5},
  year={2026},
  organization={IEEE}
}

@inproceedings{francois2021metamorphic,
  author    = {François, Anton and Gori, Pietro and Glaunès, Joan},
  title     = {Metamorphic Image Registration Using a Semi-Lagrangian Scheme},
  booktitle = {Geometric Science of Information},
  year      = {2021},
  pages     = {781--788},
  publisher = {Springer},
  doi       = {10.1007/978-3-030-80209-7_84}
}

@inproceedings{gori2015joint,
  author    = {Gori, Pietro and Colliot, Olivier and Marrakchi-Kacem, Linda
               and Worbe, Yulia and Routier, Alexandre and Poupon, Cyril
               and Hartmann, Andreas and Ayache, Nicholas and Durrleman, Stanley},
  title     = {Joint Morphometry of Fiber Tracts and Gray Matter Structures
               Using Double Diffeomorphisms},
  booktitle = {Information Processing in Medical Imaging},
  series    = {Lecture Notes in Computer Science},
  volume    = {9123},
  pages     = {275--287},
  year      = {2015},
  publisher = {Springer},
  doi       = {10.1007/978-3-319-19992-4_21}
}

@inproceedings{francois2022weighted,
  author    = {François, Anton and Maillard, Matthis and Oppenheim, Catherine
               and Pallud, Johan and Bloch, Isabelle and Gori, Pietro
               and Glaunès, Joan},
  title     = {Weighted Metamorphosis for Registration of Images
               with Different Topologies},
  booktitle = {Biomedical Image Registration},
  year      = {2022},
  pages     = {8--17},
  publisher = {Springer},
  doi       = {10.1007/978-3-031-11203-4_2}
}

@inproceedings{maillard2022deep,
  author    = {Maillard, Matthis and Francois, Anton and Glaunes, Joan
               and Bloch, Isabelle and Gori, Pietro},
  title     = {A Deep Residual Learning Implementation of Metamorphosis},
  booktitle = {2022 IEEE 19th International Symposium on Biomedical Imaging (ISBI)},
  year      = {2022},
  pages     = {1--4},
  publisher = {IEEE},
  doi       = {10.1109/ISBI52829.2022.9761422}
}

@ARTICLE{8307447,
  author={Gori, Pietro and Colliot, Olivier and Kacem, Linda Marrakchi and Worbe, Yulia and Routier, Alexandre and Poupon, Cyril and Hartmann, Andreas and Ayache, Nicholas and Durrleman, Stanley},
  journal={IEEE Transactions on Medical Imaging}, 
  title={Double Diffeomorphism: Combining Morphometry and Structural Connectivity Analysis}, 
  year={2018},
  volume={37},
  number={9},
  pages={2033-2043}
  }

@incollection{charon2019fidelity,
  author    = {Charon, Nicolas and Charlier, Benjamin and Glaunes, Joan
               and Gori, Pietro and Roussillon, Pierre},
  title     = {Fidelity Metrics between Curves and Surfaces:
               Currents, Varifolds, and Normal Cycles},
  booktitle = {Riemannian Geometric Statistics in Medical Image Analysis},
  year      = {2019},
  pages     = {441--477},
  publisher = {Elsevier},
  doi       = {10.1016/B978-0-12-814725-2.00021-2}
}

@inproceedings{gori2013bayesian,
  author    = {Gori, Pietro and Colliot, Olivier and Worbe, Yulia
               and Marrakchi-Kacem, Linda and Lecomte, Sophie
               and Poupon, Cyril and Hartmann, Andreas
               and Ayache, Nicholas and Durrleman, Stanley},
  title     = {Bayesian Atlas Estimation for the Variability Analysis
               of Shape Complexes},
  booktitle = {Medical Image Computing and Computer-Assisted Intervention
               -- MICCAI 2013},
  series    = {Lecture Notes in Computer Science},
  volume    = {8149},
  pages     = {267--274},
  year      = {2013},
  publisher = {Springer},
  doi       = {10.1007/978-3-642-40811-3_34}
}

@article{gori2017bayesian,
  author    = {Gori, Pietro and Colliot, Olivier and Marrakchi-Kacem, Linda
               and Worbe, Yulia and Poupon, Cyril and Hartmann, Andreas
               and Ayache, Nicholas and Durrleman, Stanley},
  title     = {A Bayesian Framework for Joint Morphometry of Surface
               and Curve Meshes in Multi-Object Complexes},
  journal   = {Medical Image Analysis},
  volume    = {35},
  pages     = {458--474},
  year      = {2017},
  doi       = {10.1016/j.media.2016.08.011}
}

@article{lorenzi2013efficient,
  author  = {Lorenzi, Marco and Pennec, Xavier},
  title   = {Efficient Parallel Transport of Deformations in Time Series
             of Images: A Robust Framework for Longitudinal Analysis},
  journal = {IEEE Transactions on Medical Imaging},
  year    = {2013}
}

@inproceedings{durrleman2009spatiotemporal,
  author    = {Durrleman, Stanley and Pennec, Xavier and Trouv{\'e}, Alain
               and Gerig, Guido and Ayache, Nicholas},
  title     = {Spatiotemporal Atlas Estimation for Developmental Delay Detection
               in Longitudinal Datasets},
  booktitle = {Medical Image Computing and Computer-Assisted Intervention
               -- MICCAI 2009},
  series    = {Lecture Notes in Computer Science},
  volume    = {5761},
  pages     = {297--304},
  year      = {2009},
  publisher = {Springer},
  doi       = {10.1007/978-3-642-04268-3_37}
}

@article{avants2008symmetric,
  author  = {Avants, Brian B. and Epstein, Charles L. and Grossman, Murray
             and Gee, James C.},
  title   = {Symmetric Diffeomorphic Image Registration with Cross-Correlation:
             Evaluating Automated Labeling of Elderly and Neurodegenerative Brain},
  journal = {Medical Image Analysis},
  volume  = {12},
  number  = {1},
  pages   = {26--41},
  year    = {2008},
  doi     = {10.1016/j.media.2007.06.004}
}

@article{ashburner2011diffeomorphic,
  author  = {Ashburner, John},
  title   = {Diffeomorphic Registration Using Geodesic Shooting and
             Gauss-Newton Optimisation},
  journal = {NeuroImage},
  volume  = {55},
  number  = {3},
  pages   = {954--967},
  year    = {2011},
  doi     = {10.1016/j.neuroimage.2010.12.049}
}

@article{zanello2021automated,
  author  = {Zanello, Marc and Carron, Romain and Peeters, Sophie
             and Gori, Pietro and Roux, Alexandre and Bloch, Isabelle
             and Oppenheim, Catherine and Pallud, Johan},
  title   = {Automated neurosurgical stereotactic planning for intraoperative use:
             a comprehensive review of the literature and perspectives},
  journal = {Neurosurgical Review},
  volume  = {44},
  number  = {2},
  pages   = {867--888},
  year    = {2021},
  doi     = {10.1007/s10143-020-01315-1}
}

@article{pio2026imaging,
  author  = {Pio, Luca and Kassir, Rani and La Barbera, Giammarco
             and Lozach, Cecile and Bonnot, Enzo and Isla, Thomas
             and de la Plata Alcalde, Juan Pablo and Gori, Pietro
             and Bloch, Isabelle and Sarnacki, Sabine},
  title   = {3D imaging contribution in pediatric surgical oncology:
             a multi-stakeholder assessment study},
  journal = {Scientific Reports},
  volume  = {16},
  pages   = {14264},
  year    = {2026},
  doi     = {10.1038/s41598-026-44543-z},
  publisher = {Nature Publishing Group}
}

\end{document}